\documentclass{article}

\PassOptionsToPackage{numbers,compress}{natbib}
\usepackage[main, final]{neurips_2026}

\usepackage[utf8]{inputenc}   
\usepackage[T1]{fontenc}      
\usepackage{hyperref}         
\usepackage{url}              
\usepackage{booktabs}         
\usepackage{amsfonts}         
\usepackage{nicefrac}         
\usepackage{microtype}        

\usepackage[table]{xcolor}    
\definecolor{blockrow}{HTML}{E3ECF9} 
\usepackage{amsmath}
\usepackage{amssymb}
\usepackage{mathtools}
\usepackage{amsthm}
\usepackage{multirow}
\usepackage{algorithm}   
\usepackage{algorithmic} 

\providecommand{\Call}[2]{\textsc{#1}(#2)}
\usepackage[capitalize,noabbrev]{cleveref}
\usepackage{tcolorbox}
\usepackage{graphicx}
\usepackage{subcaption}

\usepackage[disable,textsize=tiny]{todonotes}
\usepackage{xspace}

\newcommand{\Ours}{\texttt{Hi-Q}\xspace}

\theoremstyle{plain}

\theoremstyle{definition}

\theoremstyle{remark}

\title{Hi-Q: Hierarchical Evidence-guided Query Refinement for Multi-Hop Question Answering}

\author{%
  Jueun Kim$^{1}$ \quad Sungho Park$^{2}$ \quad Wook-Shin Han$^{2}$\thanks{Corresponding author.} \\
  $^{1}$Department of Computer Science and Engineering, POSTECH \\
  $^{2}$Graduate School of Artificial Intelligence, POSTECH \\
  \texttt{\{jekim,shpark,wshan\}@dblab.postech.ac.kr}
}

\begin{document}

\maketitle

\begin{abstract}
A central bottleneck in multi-hop Question Answering (QA) is that the granularity at which a question is expressed often differs from the granularity at which corpus evidence is retrievable. Existing methods address this mismatch by imposing fixed graph structures over the corpus, by iteratively reformulating the query, or by executing a generated program over it, but these strategies do not explicitly decide when a query unit is already supported by evidence and when it should be refined. We formulate this bottleneck as retrievable granularity discovery and introduce \Ours, an evidence-conditioned framework for hierarchical query refinement. At each query node, a resolution operator tests whether retrieved evidence supports the current query unit; resolved nodes terminate, while unresolved nodes are expanded by a dependency-preserving binary operator and checked by a semantic coverage verifier. \Ours therefore grows a query tree whose topology is determined by corpus support signals rather than by a fixed decomposition template or a pre-built graph. We evaluate \Ours on three multi-hop QA benchmarks over their complete corpora, where dependent evidence must be found among up to millions of open-domain passages. There it reaches 52.3 EM and 64.0 F1 on average, ahead of the iterative retrieval baseline IRCoT by 15.2 EM / 18.2 F1 and of the graph-based RAG baseline PropRAG by 11.6 EM / 12.0 F1 on MuSiQue, without corpus-wide graph construction. On the smaller controlled corpora used to compare graph-based methods, \Ours also attains the best accuracy, 57.9 EM and 69.3 F1 on average, ahead of PropRAG by 5.6 EM / 3.9 F1 and IRCoT by 13.7 EM / 15.8 F1. The project page is available at \url{https://hi-q-project.github.io/}.
\end{abstract}
\section{Introduction}
\label{sec:introduction}

\begin{figure}[t]
  \centering
  \includegraphics[width=\linewidth]{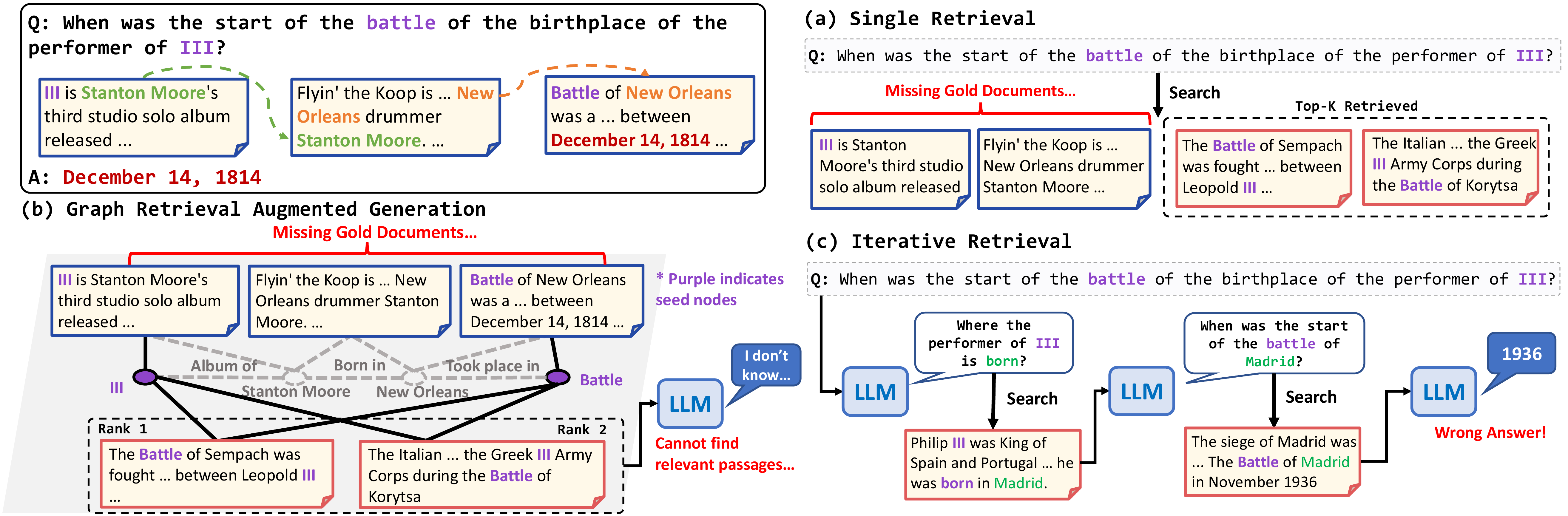}
  \caption{
\textbf{Granularity failures in multi-hop retrieval.}
(a) Single retrieval, (b) Graph RAG, and (c) iterative retrieval on the same query and corpus.
}
  \label{fig:motivation}
\vspace{-7mm}
\end{figure}

Multi-hop Question Answering (QA) requires resolving multiple interdependent reasoning steps embedded within a single natural language query.
Consider the query in Figure~\ref{fig:motivation}: ``When was the start of the battle of the birthplace of the performer of III?''
Answering this question requires first identifying the performer of ``III,'' then determining that person's birthplace, and finally finding the start date of the battle associated with that location.
Thus, a single query implicitly compresses a chain of dependent informational needs into one surface form.

A central difficulty in multi-hop QA is that the unit at which a question can be \emph{logically expressed} is often different from the unit at which evidence can be \emph{reliably retrieved}.
The facts needed to answer a multi-hop question are typically distributed across documents at a fine-grained level, while the input query presents them as a single coarse-grained sentence.
As a result, even if a question admits a plausible logical decomposition, it remains unclear \emph{a priori} which intermediate formulation is best aligned with the corpus for retrieval.
Queries that are too coarse entangle multiple reasoning aspects and cause retrieval interference, while queries that are too fine may lose contextual constraints and lead to over-decomposition.
Therefore, the key challenge is not merely how to decompose a question logically, but how to discover the \emph{retrievable granularity} at which each reasoning step becomes operationally answerable.

Figure~\ref{fig:motivation} illustrates three concrete failure modes arising from this granularity mismatch.
First, single-shot retrieval fails when a coarse query entangles several reasoning constraints: top-ranked passages may match surface terms such as ``III'' or ``battle'' without covering the full evidence chain. 
Second, Graph RAG methods add corpus-side structure, but their pre-computed graphs impose a fixed, query-agnostic granularity and may reduce reasoning to surface-level seed alignment. 
Third, iterative retrieval methods adapt the query over time, but their reformulations are not explicitly checked for evidence support, so an early wrong intermediate query can propagate through later retrieval steps. 
Together, these failures reveal the need for a control mechanism that decides which query unit is currently retrievable under the given corpus.

We introduce {\Ours}, a framework for coarse-to-fine, evidence-guided query refinement that answers these three failures with two coupled mechanisms.
Failure-aware granularity control tests whether the current query unit is already supported by retrieved evidence before refining it, so a query that single retrieval or a fixed graph already answers is never decomposed.
Dependency-preserving hierarchical decomposition then expands only the unresolved nodes, resolving prerequisite sub-queries before dependent ones so that an unchecked intermediate query cannot propagate, while a semantic coverage verifier confirms that each binary split preserves the intent of its parent.
\Ours therefore does not assume the retrieval-aligned query unit in advance, but discovers it through retrieval-and-answering feedback.

Our contributions follow this challenge-to-component mapping:

\begin{itemize}
    \item \textbf{Retrievable granularity discovery.}
    We formulate multi-hop RAG as the problem of identifying the query unit at which a reasoning step becomes both retrievable and answerable under a given corpus.

    \item \textbf{Failure-aware granularity control.}
    We propose an evidence-conditioned control policy that expands a query node only when a resolution operator detects insufficient evidence support, refining coarse queries while avoiding unnecessary decomposition of already answerable queries.
    We show that this choice is a cost-sensitive threshold on unresolved support.

    \item \textbf{Dependency-preserving hierarchical refinement.}
    We introduce a binary expansion operator that resolves prerequisite sub-queries first and propagates their results to downstream retrieval, reducing under-specified retrieval and error propagation.

    \item \textbf{Evaluation under full-corpus retrieval.}
    Across three multi-hop benchmarks, \Ours outperforms graph-based, iterative, and code-executing agent baselines without pre-built knowledge graphs or task-specific fine-tuning, and a cost-matched configuration is both cheaper and more accurate than iterative retrieval at the same number of LLM calls.
    We validate the unresolved-support signal through trigger diagnostics, ablations, and reader and embedding substitutions.
\end{itemize}

%



\section{Related Work}
\label{sec:related_work}

\paragraph{Retrieval-Augmented Generation (RAG).}
Standard RAG holds both ends fixed: the query is issued as written, and the corpus is indexed as flat passages.
It integrates external knowledge into LLMs~\cite{rag}, with dense retrievers such as DPR~\cite{dpr} and recent embedding models~\cite{contriever,nvembed} improving similarity-based evidence acquisition.
This is effective when the question and the retrievable evidence already sit at a similar granularity.
In multi-hop QA they often do not: a single coarse query matches passages on isolated surface terms while missing the complete reasoning chain~\cite{hipporag,hipporag2,proprag}.
This query--evidence granularity mismatch is the gap the remaining lines of work, and \Ours, attempt to close.

\paragraph{Graph RAG.}
Graph-based RAG moves the granularity decision to the corpus side, but makes it before any query arrives.
GraphRAG~\cite{graphrag} and RAPTOR~\cite{raptor} build hierarchical summaries, while HippoRAG~\cite{hipporag,hipporag2} and PropRAG~\cite{proprag} retrieve through graph- or proposition-level structures.
These help when the pre-computed units happen to match a question's evidence needs, but the structure is query-agnostic by construction and carries a corpus-wide pre-computation cost.
\Ours instead builds a dependency-ordered query tree online and expands only unresolved nodes, without corpus-wide graph construction.

\paragraph{Iterative Retrieval.}
Iterative retrieval issues a sequence of queries instead of one.
IRCoT~\cite{ircot}, ReAct~\cite{react}, and Self-Ask~\cite{selfask} generate each follow-up query online from intermediate findings, whereas Least-to-Most prompting~\cite{leasttomost} and Decomposed Prompting~\cite{decomposed_prompting} plan the sequence up front by decomposing the question, and TRQA~\cite{trqa} and Q-DREAM~\cite{qdream} learn such decompositions.
None of them tests whether the query it issues is resolvable from what the corpus returns: a follow-up or a split can be linguistically valid yet unnecessary, over-fragmented, or misaligned with the atomic facts the corpus exposes, and once an early step selects the wrong bridge entity or drops a constraint, later queries amplify the error.
Self-reflective and corrective RAG~\cite{selfrag,crag} do assess whether retrieved documents support the generation, but they use that judgment to filter or re-retrieve documents for a flat answer, not to decide at what granularity the next query should be issued.
\Ours instead starts from the original question and splits it step by step, testing each unit against retrieved evidence and refining only those the corpus fails to resolve, so the granularity of its queries is fitted to the corpus rather than fixed in advance.

\paragraph{Retrieval Strategy Selection.}
Retrieval strategy selection decides how retrieval is orchestrated for a question rather than how each query is expressed.
Coding agents~\cite{codingagent}, Recursive Language Models~\cite{rlm}, and PyRAG~\cite{pyrag} read the question and write a program that decides when and how to call retrieval and answering tools, while Adaptive-RAG~\cite{adaptiverag} trains a classifier of question complexity that routes each question to one of a few fixed strategies: no, single-step, or multi-step retrieval.
Both choose the strategy from the question rather than from what retrieval returns.
The route is fixed before any evidence is seen (Appendix~\ref{sec:query_routing}), and although an agent can recover from an execution failure at run time (Appendix~\ref{sec:agentic_comparison}), the recovery re-runs or patches the program rather than identifying the granularity at which a retrieval call failed and refining the query to it.
What separates \Ours is therefore the state on which control is conditioned, not the presence of learning.

\section{Method}
\label{sec:method}

\subsection{Problem Formulation}
\label{sec:formulation}

We formulate multi-hop QA as evidence-conditioned adaptive search over query granularity.
\Ours first tests whether the current query unit is supported by retrieved evidence, and expands it into smaller dependency-aware sub-queries only when the query remains unresolved.
This requires an explicit search state, because retrieval feedback determines whether a query should remain at its current coarse granularity or be refined.

\begin{figure*}[!t]
  \centering
  \includegraphics[width=\linewidth]{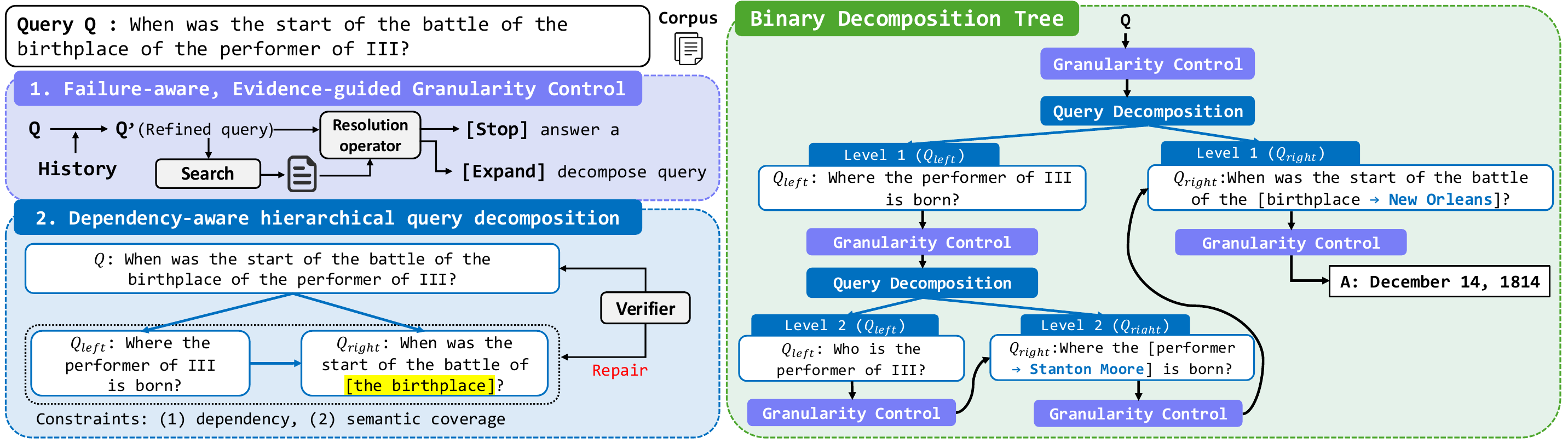}
    \caption{
    \textbf{Overview of \Ours.}
    \Ours treats multi-hop QA as evidence-conditioned search over query granularity.
At each node, the resolution operator first tests whether retrieved evidence supports the current query unit.
Resolved nodes terminate; unresolved nodes trigger dependency-ordered binary decomposition, where the prerequisite branch is resolved before the dependent branch.
A semantic coverage verifier repairs invalid splits before recursion.
The resulting binary decomposition tree is determined by corpus support signals rather than by a fixed graph or a predetermined decomposition template.
    }
  \label{fig:overview}
\end{figure*}
\vspace{-2mm}

\paragraph{Notation.}
Let $C$ denote the retrieval corpus, $Q$ the original question, and $q$ a query node in the search tree.
Let $R_k(q,C)$ be the retriever that returns the top-$k$ passages for query $q$.
We write $\mathcal{H}$ for the accumulated interaction history and $d$ for the current recursion depth.

\paragraph{Search state.}
\Ours maintains a state $x = (q, \mathcal{H}, d)$, where $q$ is the current query node, $\mathcal{H}$ is the accumulated interaction history, and $d$ is the current recursion depth.
The search objective is to find a leaf set whose queries are individually resolvable under corpus evidence and whose dependency composition preserves $Q$.

\paragraph{Resolution operator.}
Let $\mathcal{G}(q, \mathcal{H}, C) \rightarrow (s, a, D)$ be a resolution operator, where $s \in \{\textsc{resolved}, \textsc{unresolved}\}$ is the resolution status, $a$ is an answer when $s=\textsc{resolved}$, and $D = R_k(q', C)$ is the evidence retrieved for the refined query $q' = \mathrm{Refine}(q, \mathcal{H}, Q)$, which rewrites $q$ from the history while keeping it anchored to $Q$ (Section~\ref{sec:granularity_control}).
Once $\mathcal{G}$ has run, the node carries that evidence as well, giving the post-resolution state $\tilde{x} = (x, D)$.
For analysis only, we additionally assign an optional diagnostic label $\tau$ to unresolved cases; $\tau$ is not consumed by the control policy.

\paragraph{Policy.}
The policy acts on that state together with the resolution status, since whether to stop is decided by what the retrieved evidence supported:
\begin{equation*}
\pi(\tilde{x}, s) =
\begin{cases}
\textsc{stop} & \text{if } s = \textsc{resolved} \\
\textsc{fail} & \text{if } d = d_{\max} \text{ or } q \text{ is non-decomposable} \\
\textsc{expand} & \text{otherwise}
\end{cases}
\end{equation*}

\paragraph{Expansion.}
When \textsc{expand} is selected, a binary expansion operator $\mathcal{B}(q, \mathcal{H}, Q)$ proposes a pair $(q_{\text{left}}, q_{\text{right}})$ subject to two constraints: (i) a \emph{dependency constraint} $q_{\text{left}} \prec q_{\text{right}}$, meaning the prerequisite branch must be resolved before the dependent branch; (ii) a \emph{semantic coverage constraint} $\mathcal{V}(Q, q, q_{\text{left}}, q_{\text{right}})$, ensuring that resolving $q_{\text{left}}$ followed by $q_{\text{right}}$ recovers the intent of $q$ without omission or drift.
The answer to $q_{\text{left}}$ updates $\mathcal{H}$ before $q_{\text{right}}$ is resolved.

\subsection{Failure-aware, evidence-guided granularity control}
\label{sec:granularity_control}
\Ours decides whether to decompose a query from the evidence it retrieves rather than from how complex the query looks.
At each node, the resolution operator $\mathcal{G}$ first tries to resolve the query at its current granularity: it refines the query from the history $\mathcal{H}$ while keeping it anchored to $Q$, retrieves the top-$k$ passages $D$, and has a retrieval-grounded reader answer from them.
Refinement precedes retrieval because a dependent sub-query often contains a reference whose meaning is fixed only by an earlier step; once the performer of ``III'' is resolved to Stanton Moore, the query about ``the birthplace of the performer of III'' is issued as a query about the birthplace of Stanton Moore (Figure~\ref{fig:overview}).
This replaces abstract references with resolved entities, reducing retrieval interference while preserving the informational role of the current node in the original multi-hop question.
The reader's output is the control signal.
If the reader finds an answer, the retrieved evidence supports the query unit and the node is \textsc{resolved} and terminates; if it does not, the evidence does not jointly support the unit, and the node is \textsc{unresolved} and becomes eligible for expansion.

\paragraph{Threshold rule.}
The \textsc{stop}/\textsc{expand} decision rests on an estimate of whether the current node is unresolved; we derive the decision rule from the two ways the decision can fail.
A query that is too coarse entangles several constraints, so the retriever returns passages that match isolated terms and miss the full chain; decomposition then exposes smaller evidence needs that can be retrieved and verified separately.
A query that is already answerable should not be split merely because it looks complex, since an unnecessary split can drop constraints and introduce spurious intermediate goals.
Writing the two errors as costs turns the decision into a threshold.
For an expansion-admissible node in state $\tilde{x}$, let $Z \in \{\mathrm{R}, \mathrm{U}\}$ denote whether the node is in fact resolvable ($\mathrm{R}$) or unresolved ($\mathrm{U}$) from its retrieved evidence, of which the operator's status $s$ is an estimate.
Let $\Delta_{\mathrm{R}}(\tilde{x}) > 0$ be the cost-to-go penalty of expanding a resolvable node and $\Delta_{\mathrm{U}}(\tilde{x}) > 0$ that of stopping at an unresolved one, each measured over the resulting subtree and therefore including descendant retrieval and LLM calls, drift risk, synthesis, and terminal answer loss.
Minimizing conditional expected cost yields a threshold rule, derived in Appendix~\ref{sec:threshold_proof}:
\begin{equation}
\pi^{*}(\tilde{x}) = \textsc{expand}
\iff
\Pr[Z = \mathrm{U} \mid \tilde{x}] \;\geq\; \theta(\tilde{x}) := \frac{\Delta_{\mathrm{R}}(\tilde{x})}{\Delta_{\mathrm{R}}(\tilde{x}) + \Delta_{\mathrm{U}}(\tilde{x})} .
\label{eq:threshold}
\end{equation}
Because the objective includes downstream cost-to-go, this decision is node-wise but not myopic; we do not claim global optimality of the resulting query tree, and \textsc{fail} is handled separately as a budget or feasibility action.

\Ours instantiates the estimate without training, as a test that reads it off the reader's attempt to answer from the retrieved passages.
The reader is instructed to answer from the retrieved passages, extracting the answer span at the root, so \textsc{unresolved} means that the retrieved passages do not support the query, not that the model declined to answer.
This test is deliberately the simplest instantiation of the rule: it ignores depth and cost weights, which is defensible because stopping at an unresolved node forfeits the answer while an unnecessary expansion costs far less, an asymmetry the experiments measure (decomposing every node costs $3.4$ EM on the ablation sample of Section~\ref{sec:ablation}, whereas never decomposing, i.e.\ single retrieval, trails \Ours by $9.7$ EM in the controlled-corpus setting of Table~\ref{tab:full_corpus}), so $\Delta_{\mathrm{U}} \gg \Delta_{\mathrm{R}}$ places $\theta$ well below one half and favors expanding whenever support is incomplete.
Keeping the estimator this simple also attributes the gains in Section~\ref{sec:experiments} to the control structure rather than to the estimator.
Any instantiation, however, must observe the evidence state: when the same query calls for opposite actions under different retrieved evidence, a policy that sees only the query incurs a strictly positive regret that no amount of query-only training data removes, as we prove in Appendix~\ref{sec:threshold_proof}; this is what separates \Ours from methods that choose a retrieval strategy from the question alone, whose learned routing collapses to a near-constant policy in Appendix~\ref{sec:query_routing}.
The training-free test is a hard classifier $\hat{Z}$ and is not claimed to compute the posterior: provided it observes the evidence state, a calibrated classifier, an entailment model, or a trained cost-sensitive router can replace it without changing \Ours's control semantics (Appendix~\ref{sec:learned_router}), and Section~\ref{sec:experiments} evaluates how precise it is in practice.

\subsection{Dependency-aware hierarchical query decomposition}
\label{sec:decomposition}
Dependency-aware decomposition turns an unresolved query into an ordered split whose sub-goals can be tested against evidence.
\Ours splits binary rather than N-ary (multi-way) to avoid over-fragmenting the question or skipping essential intermediate bridge facts: a single multi-way split jumps to the finest granularity at once, whereas binary expansion moves from coarse to fine one step at a time, so at each step the resolution test can check whether the current granularity is already retrievable before splitting further.
Splitting in two restricts sequential depth rather than expressiveness: a plan with $m$ leaf information needs is organized as at most $m-1$ binary reductions, with the expansion operator and the verifier unchanged, and how many reductions occur is decided by the controller, since a node resolvable from its retrieved passages terminates there.
Appendix~\ref{sec:high_arity} tests this on a controlled set whose leaf needs rise from two to five: binary expansion holds its accuracy across that range, branching with unrestricted arity does not improve on it, and combining the branches in one flat $N$-ary step instead of sequential binary reductions is substantially worse.

Expansion is subject to the two constraints of Section~\ref{sec:formulation}.
The dependency constraint makes the left branch a prerequisite for the right: given an unresolved query $q$, $\mathcal{B}(q,\mathcal{H},Q)$ proposes exactly two sub-queries, $q_{\text{left}}$, which resolves the bridge fact, and $q_{\text{right}}$, which uses that bridge to approach the parent query's answer.
The split follows entity references and relational dependencies (compositional, temporal, or causal) rather than the surface syntax of the question, so the decomposition order matches the order in which retrieval must proceed.
That order has to be executed and not merely proposed, because a dependent sub-query is often under-specified until its prerequisite is resolved: \Ours resolves $q_{\text{left}}$ first, writes its answer, if any, together with its retrieved evidence and intermediate context into the history $\mathcal{H}$, and then resolves $q_{\text{right}}$ under the updated history.
For the running example, the system must identify the performer of ``III'' before it can retrieve evidence about the relevant birthplace and battle.
The dependency is enforced at the evidence-state level rather than as a brittle requirement that the left branch must always produce a final answer: even if $q_{\text{left}}$ remains unresolved, its retrieved evidence and partial context are retained in $\mathcal{H}$ to ground the dependent branch.

The semantic coverage constraint is enforced by the verifier $\mathcal{V}$ and prevents local decomposition errors from becoming downstream reasoning errors.
Before the two sub-queries are solved, $\mathcal{V}$ checks whether resolving $q_{\text{left}}$ followed by $q_{\text{right}}$ would recover the intent of $q$ without adding, omitting, or reordering essential constraints.
If the split is inconsistent, $\mathcal{V}$ repairs the two sub-queries before they enter the recursive solver; we allow at most one repair attempt, and if the revised split is still inconsistent, the node is marked non-decomposable and decomposition stops for that branch.
This guardrail is especially important at deeper recursion depths, where a small semantic drift in one split can compound across later branches.

Recursive expansion is bounded so that refinement remains a controlled search rather than an open-ended reasoning loop.
If a sub-query is resolved, recursion stops at that node.
If a sub-query is non-decomposable or the maximum depth is reached, the policy takes \textsc{fail}: the branch terminates without an answer, and the solver proceeds with the evidence gathered so far.
After both branches return, a synthesis step aggregates intermediate answers, retrieved evidence, unresolved sub-goals, and history into a final response consistent with the original query $Q$. When multiple evidence-supported candidates are retained from earlier resolution steps, synthesis disambiguates them using consistency with other sub-query results rather than committing to the highest-ranked passage alone.
Thus, \Ours constructs a finite dependency-ordered binary tree whose leaves correspond to the evidence-aligned query units used for final reasoning.
The full recursive procedure is presented in Algorithm~\ref{alg:ours} (Appendix~\ref{sec:algorithm}), and the prompt templates are listed in Appendix~\ref{sec:Prompts}.
\section{Experiments}
\label{sec:experiments}
\subsection{Experimental Setup}
\paragraph{Datasets and evaluation protocol.}
We evaluate \Ours on HotpotQA~\cite{hotpotqa}, MuSiQue~\cite{musique}, and 2WikiMultiHopQA~\cite{2wiki}.
Following prior graph-based RAG methods~\cite{hipporag,hipporag2,proprag}, we sample 1{,}000 validation questions per dataset and evaluate each question under two retrieval regimes.
The first, the \emph{full-corpus} setting, is the benchmark's complete corpus, in which retrieval scales to 5{,}233{,}235 passages for HotpotQA, 139{,}416 for MuSiQue, and 430{,}225 for 2WikiMultiHopQA.
The second, the \emph{controlled-corpus} setting, is the corpus used by these graph-based methods, pooled from the supporting and distractor passages of the sampled questions, with 9{,}811 passages for HotpotQA, 11{,}656 for MuSiQue, and 6{,}119 for 2WikiMultiHopQA.
EM and token-level F1 measure final QA accuracy; Recall@$k$ and its pooling protocol for multi-query methods are defined in Appendix~\ref{sec:eval_details}.
Appendix~\ref{sec:strict_eval} repeats the evaluation under strict document-level grounding.
Unless stated otherwise, differences between methods are tested with a paired question-level bootstrap over the per-question predictions behind each table, using $10{,}000$ resamples and $95\%$ percentile intervals.

\paragraph{Baselines and implementation.}
We compare against four families of methods, the first three of which cover the three failure modes in Figure~\ref{fig:motivation}: single retrieval tests whether one coarse query suffices, graph-based methods test corpus-side structuring before query-specific reasoning, and iterative methods test predefined or model-driven follow-up query generation.
Single retrieval issues the original question once to a dense retriever, represented by NV-Embed-v2~\cite{nvembed}.
Graph-based RAG structures the corpus before any question arrives, through hierarchical summaries, entity graphs, or proposition paths, represented by RAPTOR, GraphRAG, HippoRAG, HippoRAG~2, and PropRAG~\cite{raptor,graphrag,hipporag,hipporag2,proprag}.
Iterative methods issue a sequence of queries, generated online from intermediate reasoning or planned up front by decomposing the question, represented by IRCoT, Self-Ask, and Least-to-Most~\cite{ircot,selfask,leasttomost}.
A comparison with the fine-tuned decomposer TRQA under its own protocol is in Appendix~\ref{sec:decomp_comparison}.
Agentic methods cast multi-hop QA as code generation and execution over the corpus, represented by a coding agent~\cite{codingagent} and a Recursive Language Model~\cite{rlm}.
In the full-corpus setting, PropRAG is omitted for HotpotQA-full, because running its LLM-based graph construction over 5.2M passages would cost more than \$2{,}500 in API calls alone, beyond the budget available to us.
All methods use GPT-4o-mini as the reader unless otherwise stated, and \Ours uses NV-Embed-v2, $k=5$, maximum recursion depth $d_{\max}=4$, and temperature $0$.
Appendices~\ref{sec:open_source_results} and~\ref{sec:embedding_robustness} measure the robustness of these two choices by substituting other readers (Llama-3.3-70B, Qwen3-30B-A3B) and other embedding models (text-embedding-3-large, Qwen3-Embedding-8B).

\subsection{Main Results}
\paragraph{\Ours outperforms iterative, graph-based, and agentic baselines.}
Table~\ref{tab:full_corpus} evaluates the 1{,}000-question subsets in the full-corpus setting.
\Ours reaches 52.3 EM and 64.0 F1 on average, outperforming IRCoT by 15.2 EM / 18.2 F1, with per-benchmark margins of 5.4 EM / 6.3 F1 on HotpotQA-full, 13.0 EM / 15.7 F1 on MuSiQue-full, and 27.0 EM / 32.5 F1 on 2Wiki-full.
It also surpasses PropRAG by 11.6 EM / 12.0 F1 on MuSiQue-full and by 9.9 EM / 12.1 F1 on 2Wiki-full, the two corpora over which PropRAG's graph could be constructed within our compute budget, and does so without constructing a graph over the corpus.
\Ours also improves over the coding agent and the Recursive Language Model by $30.8$ EM / $26.9$ F1 and by $24.4$ EM / $26.7$ F1 on average.
An agent's program controls when retrieval is called, but the query passed to each call is still written by the model with no check that it is expressed at a granularity the corpus can match, so a coarse or under-specified query returns the same distractors inside a program as in single-shot retrieval.
\Ours makes that check the control signal, expanding a node only when its retrieved evidence fails to resolve it and rewriting each dependent sub-query with the resolved bridge entity before issuing it.

\paragraph{The gains also hold in the controlled-corpus setting of prior work.}
To compare against a wider range of graph-based methods, we also evaluate in the controlled-corpus setting used in the HippoRAG and PropRAG evaluations~\cite{hipporag,hipporag2,proprag} (Table~\ref{tab:full_corpus}, Controlled corpus).
\Ours achieves 57.9 EM and 69.3 F1 on average, above every graph-based method, including HippoRAG~2 by 7.3 EM / 5.9 F1 and PropRAG by 5.6 EM / 3.9 F1, and above IRCoT by 13.7 EM / 15.8 F1.
The largest gains again occur on MuSiQue, where shortcut reasoning is limited and evidence dependencies must be resolved explicitly: 7.3 EM / 5.0 F1 over PropRAG and 13.1 EM / 14.8 F1 over IRCoT.
The gold passages are present in both settings, but here they compete with far fewer candidates, which compresses the differences that the full corpora expose.
Every improvement over IRCoT and PropRAG is significant, in all six controlled-corpus comparisons and all five full-corpus ones, with $19$ of the $22$ EM and F1 tests at $p < 10^{-4}$.
The smallest margin, $3.2$ F1 over PropRAG on 2Wiki in the controlled-corpus setting, reaches $p = 0.0033$, and the F1 gain over PropRAG on MuSiQue-full has a $95\%$ confidence interval of $[+9.44, +14.57]$.

\paragraph{Higher pooled recall does not by itself produce better answers.}
Self-Ask attains the highest average Recall@5 in the controlled-corpus setting, but it issues more sub-queries and retrieves roughly 33\% more documents per question than \Ours, and because $k$ is applied per retrieval call while recall is computed over the union of all returned passages, more calls buy a larger effective budget at the same nominal $k$.
A larger candidate pool therefore raises recall mechanically while handing the reader more distractors, and the pooled score says nothing about execution alignment: whether each required passage was visible at the reasoning step that needed it, and whether the information it carried reached final synthesis.
Self-Ask accordingly reaches higher average recall with much lower answer accuracy, 35.7 EM against \Ours's 57.9.
\Ours's advantage is therefore not raw retrieval volume; it targets query units whose retrieved evidence is also resolvable by the reader.

\begin{table*}[ht]
\caption{\textbf{Main results.} QA and retrieval performance in the full-corpus setting, over each benchmark's complete corpus (Full corpus), and in the controlled-corpus setting of prior work, over pooled supporting and distractor passages (Controlled corpus).
$^{*}$Results are reproduced.
$^{\dagger}$PropRAG is reported only on MuSiQue and 2Wiki under full-corpus retrieval.}
\label{tab:full_corpus}
\centering
\footnotesize
\setlength{\tabcolsep}{3pt}
\renewcommand{\arraystretch}{0.95}
\resizebox{\textwidth}{!}{%
\begin{tabular}{l|cccc|cccc|cccc|cccc}
\toprule
\multirow{2}{*}{Method}
& \multicolumn{4}{c|}{HotpotQA}
& \multicolumn{4}{c|}{MuSiQue}
& \multicolumn{4}{c|}{2Wiki}
& \multicolumn{4}{c}{Avg}\\
\cmidrule(lr){2-5} \cmidrule(lr){6-9} \cmidrule(lr){10-13} \cmidrule(lr){14-17}
& EM & F1 & R@2 & R@5
& EM & F1 & R@2 & R@5
& EM & F1 & R@2 & R@5
& EM & F1 & R@2 & R@5 \\
\midrule
\rowcolor{blockrow}\multicolumn{17}{l}{\textit{Full corpus}} \\
PropRAG$^{*,\dagger}$  & \multicolumn{4}{c|}{N/A} & 25.8 & 38.6 & 44.4 & 57.9 & 52.1 & 59.3 & 56.3 & 74.7 & \multicolumn{4}{c}{N/A} \\
\midrule
Self-Ask$^{*}$         & 30.0 & 37.5 & 63.2 & 74.5 & 22.2 & 28.1 & 49.9 & 65.2 & 41.7 & 46.9 & \textbf{62.0} & \textbf{89.9} & 31.3 & 37.5 & 58.4 & 76.5 \\
Least-to-Most$^{*}$    & 45.4 & 59.1 & 58.8 & 71.0 & 24.3 & 34.0 & 44.4 & 56.3 & 30.7 & 34.0 & 57.8 & 68.2 & 33.5 & 42.4 & 53.7 & 65.2 \\
IRCoT$^{*}$            & 52.0 & 63.6 & 68.9 & 80.2 & 24.4 & 34.9 & 46.9 & 63.0 & 35.0 & 38.9 & 57.5 & 76.9 & 37.1 & 45.8 & 57.8 & 73.4 \\
\midrule
Coding agent$^{*}$     & 31.1 & 47.1 & 64.9 & 76.6 & 16.8 & 29.6 & 43.7 & 58.4 & 16.5 & 34.7 & 59.9 & 81.1 & 21.5 & 37.1 & 56.2 & 72.0 \\
RLM$^{*}$              & 30.1 & 41.1 & 21.5 & 25.8 & 19.6 & 29.1 & 21.7 & 29.7 & 34.0 & 41.6 & 25.9 & 35.4 & 27.9 & 37.3 & 23.0 & 30.3 \\
\midrule
\textbf{\Ours}         & \textbf{57.4} & \textbf{69.9} & \textbf{70.5} & \textbf{82.0} & \textbf{37.4} & \textbf{50.6} & \textbf{53.4} & \textbf{70.8} & \textbf{62.0} & \textbf{71.4} & 61.0 & 86.3 & \textbf{52.3} & \textbf{64.0} & \textbf{61.6} & \textbf{79.7} \\
\midrule
\rowcolor{blockrow}\multicolumn{17}{l}{\textit{Controlled corpus}} \\
NV-Embed-v2            & 57.3 & 71.0 & 84.1 & 94.5 & 32.8 & 46.0 & 52.7 & 69.7 & 54.4 & 60.8 & 67.1 & 76.5 & 48.2 & 59.3 & 68.0 & 80.2 \\
\midrule
RAPTOR                 & 50.6 & 64.7 & 78.6 & 90.2 & 27.7 & 39.2 & 49.1 & 61.0 & 39.7 & 48.4 & 58.4 & 66.0 & 39.3 & 50.8 & 62.0 & 72.4 \\
GraphRAG               & 51.4 & 67.6 & -- & -- & 27.0 & 42.0 & -- & -- & 45.7 & 61.0 & -- & -- & 41.4 & 56.9 & -- & -- \\
HippoRAG               & 46.3 & 60.0 & 60.1 & 78.5 & 24.0 & 35.9 & 41.8 & 52.4 & 59.4 & 67.3 & 68.4 & 87.0 & 43.2 & 54.4 & 56.8 & 72.6 \\
HippoRAG 2             & 56.3 & 71.1 & 80.5 & 95.7 & 35.0 & 49.3 & 53.5 & 74.2 & 60.5 & 69.7 & 74.6 & 90.2 & 50.6 & 63.4 & 69.5 & 86.7 \\
PropRAG$^{*}$          & 59.1 & 73.8 & 85.1 & \textbf{96.9} & 37.7 & 52.2 & 55.2 & 75.8 & 60.2 & 70.2 & 74.8 & 91.3 & 52.3 & 65.4 & 71.7 & 88.0 \\
\midrule
Self-Ask$^{*}$         & 45.4 & 62.3 & 82.5 & 92.6 & 29.2 & 43.8 & \textbf{60.0} & \textbf{77.4} & 32.6 & 58.7 & \textbf{84.4} & \textbf{98.8} & 35.7 & 54.9 & \textbf{75.6} & \textbf{89.6} \\
Least-to-Most$^{*}$    & 55.9 & 71.3 & 78.5 & 92.4 & 29.1 & 39.5 & 51.8 & 66.2 & 37.0 & 41.5 & 68.0 & 75.2 & 40.7 & 50.8 & 66.1 & 77.9 \\
IRCoT$^{*}$            & 59.9 & 72.0 & 85.2 & 95.2 & 31.9 & 42.4 & 55.9 & 74.0 & 40.9 & 46.0 & 76.8 & 90.8 & 44.2 & 53.5 & 72.6 & 86.7 \\
\midrule
\textbf{\Ours}         & \textbf{64.3} & \textbf{77.4} & \textbf{85.8} & 96.2 & \textbf{45.0} & \textbf{57.2} & 56.9 & 75.7 & \textbf{64.5} & \textbf{73.4} & 78.5 & 93.4 & \textbf{57.9} & \textbf{69.3} & 73.7 & 88.4 \\
\bottomrule
\end{tabular}%
}
\vspace{-2mm}
\end{table*}

\subsection{The Unresolved-Support Signal}

\begin{figure}[h]
\centering
\includegraphics[width=\linewidth]{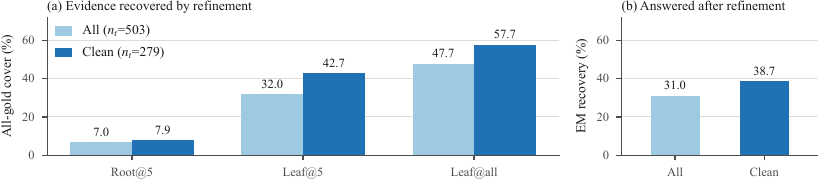}
\caption{Trigger recovery on MuSiQue. (a) All-gold cover of Root@5, Leaf@5, and Leaf@\textit{all} on all triggered questions ($n_t{=}503$) and on the Clean subset ($n_t{=}279$). (b) EM recovery, the fraction of triggered questions ultimately answered correctly.}
\label{fig:trigger-recovery}
\vspace{-2mm}
\end{figure}

\paragraph{The signal is actionable: refinement recovers evidence that root retrieval misses.}
Decomposition makes retrievable what the coarse root query could not retrieve.
A trigger fires when the passages retrieved for the root query fail to resolve the question, so triggered cases are those in which the question at its original granularity did not retrieve its evidence.
On the triggered MuSiQue cases, we test whether the sub-queries issued after decomposition retrieve the gold passages that the root query missed, comparing root@5 with \textbf{Leaf@5}, the top-5 of the pooled leaf retrievals ranked by retriever similarity and therefore budget-matched to root@5, and with \textbf{Leaf@\textit{all}}, the full pool.
Figure~\ref{fig:trigger-recovery} reports all-gold cover, the fraction of triggered questions for which every gold passage is retrieved.
On the Clean subset, the root query covers all gold passages in only $7.9\%$ of triggered questions; the sub-queries raise this to $42.7\%$ at the same five-passage budget and to $57.7\%$ over the full pool, and $38.7\%$ of these initially failed questions are ultimately answered correctly.
Because most of the gain is already present at the matched Leaf@5 budget, it comes from the sub-queries ranking the gold passages highly rather than from retrieving more passages: the evidence was in the corpus all along, and refining the query to the granularity at which the corpus stores it is what makes it retrievable.

\begin{figure}[h]
\centering
\includegraphics[width=\linewidth]{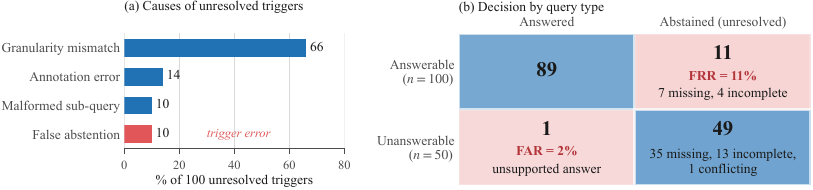}
\caption{Trigger analysis on MuSiQue. (a) Causes of 100 sampled \textsc{unresolved} triggers. (b) Decisions on 100 answerable and 50 adversarially unanswerable queries; the off-diagonal cells are the two error types, false rejection (FRR) and false acceptance (FAR), and the \textsc{unresolved} cells list the failure-type labels.}
\label{fig:trigger-analysis}
\end{figure}

\paragraph{The signal is precise enough to route decomposition without a learned classifier.}
\Ours uses a deterministic trigger: a node is expanded only when the resolution operator returns \textsc{unresolved}.
What matters is not why each unresolved-support signal arose, but how often one is raised while the retrieved evidence does support the current query, since that is the only case in which withholding an answer is the wrong action.
To test this, we manually analyze 100 randomly sampled \textsc{unresolved} triggers on MuSiQue and group them by cause in Figure~\ref{fig:trigger-analysis}(a).
Annotation errors ($14\%$) reflect defective benchmark evidence and malformed sub-queries ($10\%$) are upstream formulation failures; in both the evidence genuinely fails to support the query, so abstaining is the correct action and neither counts as a trigger error.
Of the remaining 76 triggers, 66 ($87\%$) are granularity mismatches, the case the trigger is designed to catch, and only 10 are false abstentions, in which the operator returned \textsc{unresolved} although the retrieved passages did support the query.
These 10 are the only false triggers: the signal fires on a genuinely unresolved query in $90\%$ of cases, precise enough to route decomposition without a learned classifier.

\paragraph{The signal is conservative under both answerable and adversarially unanswerable inputs.}
The resolution operator can err in two directions.
It may return \textsc{unresolved} although the retrieved evidence supports the current query, in which case a question that was already answerable at its current granularity is decomposed; or it may answer although the evidence does not support the query, in which case the node is never expanded and an unsupported answer propagates.
We measure both directions on MuSiQue: the False Rejection Rate (FRR) on 100 normal queries, and the False Acceptance Rate (FAR) on 50 adversarially unanswerable queries constructed by replacing answerable conditions following the SQuAD~2.0 methodology~\cite{squad2}; for diagnosis only, each unresolved output carries a failure-type label that the policy does not use.
Figure~\ref{fig:trigger-analysis}(b) shows FRR $=11\%$ on normal queries, with all 11 null outputs labeled as missing or incomplete evidence rather than blind abstention.
On adversarially unanswerable queries, \Ours identifies 49 of 50 as unanswerable, yielding FAR $=2\%$.
The trigger therefore rarely answers without evidence, and its errors fall on the withholding side, whose cost is an unnecessary expansion bounded by the always-decompose ablation in Table~\ref{tab:ablation_all}, rather than on the answering side, which would let an unsupported answer stand.

\subsection{Component Ablations and Reasoning Depth}
\label{sec:ablation}

\paragraph{Each part of the control loop earns its cost.}
Table~\ref{tab:ablation_all} removes the components of \Ours one at a time.
Replacing hierarchical refinement with a static one-shot decomposition lowers the average from $57.9$ EM / $69.3$ F1 to $51.5$ EM / $63.7$ F1, so retrieval feedback is needed to adjust granularity.
Removing dependency awareness causes a larger drop, to $47.1$ EM / $56.0$ F1, because a dependent sub-query issued before its prerequisite is resolved is under-specified.
Finally, replacing the failure-aware trigger with an always-decompose policy reduces performance from $60.7$ EM / $70.8$ F1 to $57.3$ EM / $68.5$ F1 on the sample of 100 queries per dataset.
This shows that unconditional decomposition fragments already answerable queries, while \Ours's trigger expands only when evidence support is insufficient.
Because always-decompose is the limiting case of over-triggering, firing on every query, this $3.4$-point drop also bounds what any trigger error can cost.
Round-trip verification has a small overall effect but matters at depth: the gap from removing it widens with the number of hops (Table~\ref{tab:verification-ablation}), reaching $1.0$ F1 at 4 hops on MuSiQue, where a decomposition error can compound across successive splits.
The verifier repairs $14.8$--$18.8\%$ of triggered decompositions, with per-dataset trigger and repair frequencies in Appendix~\ref{sec:trigger_rate}.
Consistently, generating several root decompositions and voting over them raises accuracy only modestly at near-linear token cost, indicating that early decomposition errors rarely survive to the final answer (Appendix~\ref{sec:multi_decomp}).
Each component thus removes a different failure: refinement lets granularity follow what retrieval returns, dependency ordering keeps dependent sub-queries specified, the trigger spares already answerable queries at a fraction of the cost of decomposing every node, and the verifier keeps decomposition errors from compounding at depth.

\begin{table}[ht]
\caption{Control-flow ablation of \Ours.
Bold indicates the best value in each column.
$^{\dagger}$100 randomly sampled queries per dataset (300 in total), due to the prohibitive cost of recursive always-decompose.}
\label{tab:ablation_all}
\centering
\footnotesize
\setlength{\tabcolsep}{4pt}
\renewcommand{\arraystretch}{0.95}
\begin{tabular}{l|cc|cc|cc|cc}
\toprule
\multirow{2}{*}{Setting}
& \multicolumn{2}{c|}{HotpotQA}
& \multicolumn{2}{c|}{MuSiQue}
& \multicolumn{2}{c|}{2Wiki}
& \multicolumn{2}{c}{Avg} \\
\cmidrule(lr){2-3} \cmidrule(lr){4-5} \cmidrule(lr){6-7} \cmidrule(lr){8-9}
& EM & F1 & EM & F1 & EM & F1 & EM & F1 \\
\midrule
\multicolumn{9}{l}{\textit{Decomposition variant (full subset):}} \\
\textbf{\Ours}                             & \textbf{64.3} & \textbf{77.4} & \textbf{45.0} & \textbf{57.2} & \textbf{64.5} & \textbf{73.4} & \textbf{57.9} & \textbf{69.3} \\
\quad w/o Hierarchical Decomp.\            & 58.0 & 73.6 & 39.0 & 51.2 & 57.6 & 66.3 & 51.5 & 63.7 \\
\quad w/o Dependency-aware Decomp.\        & 62.5 & 74.9 & 34.2 & 43.2 & 44.5 & 50.0 & 47.1 & 56.0 \\
\midrule
\multicolumn{9}{l}{\textit{Triggering policy ($N{=}100$ per dataset$^{\dagger}$):}} \\
\textbf{\Ours} (Failure-aware)             & \textbf{62.0} & \textbf{76.6} & \textbf{49.0} & \textbf{57.2} & \textbf{71.0} & \textbf{78.7} & \textbf{60.7} & \textbf{70.8} \\
\quad Always-decompose                     & 58.0 & 75.3 & 45.0 & 53.4 & 69.0 & 76.9 & 57.3 & 68.5 \\
\bottomrule
\end{tabular}
\end{table}

\paragraph{\Ours degrades more gracefully than baselines as reasoning depth increases.}
Figure~\ref{fig:hop_analysis} stratifies MuSiQue performance by hop count.
Every method loses accuracy as the number of hops grows, but \Ours keeps the highest F1 at each depth, reaching $55.2$ and $39.2$ F1 at 3 and 4 hops against $43.4$ and $26.9$ for IRCoT and $43.1$ and $35.4$ for PropRAG.
Deeper questions are where an early wrong bridge entity or a dropped constraint propagates furthest, so this is where resolving prerequisite sub-queries before dependent ones is expected to help, as Section~\ref{sec:decomposition} argues.

\begin{table}[htbp]
\vspace{-4mm}
\centering
\begin{minipage}[c]{0.46\linewidth}
\centering
\caption{Verification ablation by hop count (F1 on MuSiQue).}
\label{tab:verification-ablation}
\footnotesize
\setlength{\tabcolsep}{5pt}
\renewcommand{\arraystretch}{1.35}
\begin{tabular}{lcc}
\toprule
Hop & \Ours & w/o RT \\
\midrule
Overall & 57.2 & 56.5 \\
2-hop   & 64.3 & 63.7 \\
3-hop   & 55.2 & 54.4 \\
4-hop   & 39.2 & 38.2 \\
\bottomrule
\end{tabular}
\end{minipage}
\hspace{0.02\linewidth}
\begin{minipage}[c]{0.48\linewidth}
\centering
\captionof{figure}{EM and F1 on MuSiQue across hop counts.}
\label{fig:hop_analysis}
\includegraphics[width=\linewidth]{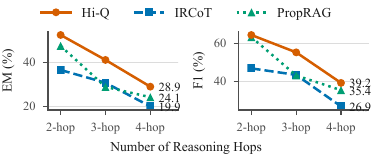}
\end{minipage}
\vspace{-4mm}
\end{table}

\subsection{Computational Cost}
\label{sec:cost}
Table~\ref{tab:cost_analysis} reports the per-question cost.
\Ours takes $13.5$ seconds against PropRAG's $23.1$, which covers graph traversal but not corpus-wide graph construction.
Against IRCoT, \Ours uses $70.5\%$ fewer tokens in total but about twice the latency and $7.5\times$ the output tokens; this overhead comes from decomposition, verification, and descendant resolution, which run only for unresolved nodes, so it is paid by the queries that fail to resolve rather than by every query.
A cost-matched configuration shortens the rationales and caps the recursion depth at $1$, leaving the policy and every operator unchanged.
At the same number of LLM calls as IRCoT ($2.93$ versus $2.92$), it uses $9.4\times$ fewer input tokens, runs $25\%$ faster, and costs $8.5\times$ less per question at GPT-4o-mini list prices (Appendix~\ref{sec:eval_details}), while scoring $10.4$ EM / $11.6$ F1 higher on average ($p \leq 0.0036$ on every benchmark).
The default configuration adds a further $3.3$ EM / $4.2$ F1 at roughly one third of IRCoT's API token cost.

\begin{table}[h]
\vspace{-4mm}
\centering
\caption{Computational cost per question, macro-averaged over the three benchmarks. Latency is end-to-end wall-clock time, retrieval and reading included; LLM calls and tokens count reader usage. Corpus embedding and PropRAG's offline graph construction are excluded.}
\label{tab:cost_analysis}
\footnotesize
\setlength{\tabcolsep}{6pt}
\renewcommand{\arraystretch}{0.95}
\begin{tabular}{lcccc}
\toprule
Method & Latency (s) & LLM Calls & Input toks & Output toks \\
\midrule
PropRAG        & 23.1 & 1    & 1{,}411  & 88  \\
IRCoT          & 6.4  & 2.92 & 43{,}701 & 74  \\
Self-Ask       & 8.0  & 6.36 & 4{,}591  & 98  \\
Least-to-Most  & 11.0 & 5.38 & 6{,}874  & 424 \\
\rowcolor{blockrow!50}\textbf{\Ours (cost-matched)} & 4.8  & 2.93 & 4{,}641  & 131 \\
\rowcolor{blockrow!50}\textbf{\Ours}                & 13.5 & 5.7  & 12{,}343 & 558 \\
\bottomrule
\end{tabular}
\end{table}

\section{Conclusion}
\label{sec:conclusion}

The main finding of this work is that multi-hop RAG benefits from treating query granularity as a control variable rather than as a fixed design choice.
\Ours operationalizes this idea by testing each query node against retrieved evidence, expanding only unresolved nodes, and preserving prerequisite-to-dependent ordering during refinement.
Across the full-corpus and controlled-corpus settings, this evidence-conditioned control improves answer accuracy over graph-based, iterative, and code-executing agent baselines while avoiding corpus-wide knowledge graph construction.
More broadly, multi-hop retrieval should be viewed not only as a problem of finding more passages, but as a problem of discovering the query granularity at which each reasoning step becomes retrievable and answerable.

\begin{ack}
This work was supported by the Institute of Information \& Communications Technology Planning \& Evaluation (IITP) grant funded by the Korea government (MSIT) (No. RS-2024-00509258, Global AI Frontier Lab, 40\%; No. IITP-2026-RS-2026-25616370, AI Star Fellowship Support Program, 20\%) and the National Research Foundation of Korea (NRF) grant funded by the Korea government (MSIT) (No. RS-2025-00517736, 40\%). This work utilized two GPUs provided by the National AI Project of the Ministry of Science and ICT (MSIT), Korea.
\end{ack}

\bibliographystyle{unsrtnat}
\bibliography{example_paper}

\newpage
\appendix
\onecolumn

\section{Algorithm}
\label{sec:algorithm}

Algorithm~\ref{alg:ours} instantiates the evidence-conditioned search policy formalized in Section~\ref{sec:formulation}.
Each line annotated with a $\pi$-action corresponds to one of the policy's three decisions (\textsc{stop}, \textsc{fail}, \textsc{expand}).
The three operators are realized as prompted-LLM modules whose prompts are listed in Appendix~\ref{sec:Prompts}:
the resolution operator $\mathcal{G}$ encapsulates query refinement, retrieval, and reference-aware answering, returning the resolution status $s \in \{\textsc{resolved}, \textsc{unresolved}\}$ together with the answer $a$ and retrieved evidence $D$;
the binary expansion operator $\mathcal{B}$ proposes a dependency-ordered pair of sub-queries;
the semantic coverage verifier $\mathcal{V}$ checks and repairs the split before recursion.
The interaction history $\mathcal{H}$ is updated in place, so the right branch is resolved under an $\mathcal{H}$ already containing the answer to the left branch.

\begin{algorithm}[h]
\caption{Evidence-conditioned hierarchical query refinement (\Ours).}
\label{alg:ours}
\begin{algorithmic}[1]
\REQUIRE Original query $Q$, corpus $C$, retrieval size $k$, max recursion depth $d_{\max}$
\ENSURE Final answer $A$
\STATE $\mathcal{H} \gets [\ ]$ \COMMENT{interaction history}
\STATE $A \gets \textsc{Solve}(Q, \mathcal{H}, 0)$
\STATE \textbf{return} $A$

\FUNCTION{\textsc{Solve}$(q, \mathcal{H}, d)$}{}
    \STATE $(s, a, D) \gets \mathcal{G}(q, \mathcal{H}, C)$ \COMMENT{resolution: refine $\to$ retrieve $\to$ answer}
    \STATE $\Call{Append}{\mathcal{H}, (q, D, a)}$
    \IF{$s = \textsc{resolved}$}{}
        \STATE \textbf{return} $a$ \COMMENT{$\pi(\tilde{x}, s) = \textsc{stop}$}
    \ENDIF
    \IF{$d = d_{\max}$}{}
        \STATE \textbf{return} $\bot$ \COMMENT{$\pi(\tilde{x}, s) = \textsc{fail}$: depth budget exhausted}
    \ENDIF

    \STATE $(q_{\text{left}}, q_{\text{right}}) \gets \mathcal{B}(q, \mathcal{H}, Q)$ \COMMENT{binary expansion}
    \IF{$(q_{\text{left}}, q_{\text{right}}) = \bot$}{}
        \STATE \textbf{return} $\bot$ \COMMENT{$\pi(\tilde{x}, s) = \textsc{fail}$: non-decomposable}
    \ENDIF
    \STATE $(q_{\text{left}}, q_{\text{right}}) \gets \mathcal{V}(Q, q, q_{\text{left}}, q_{\text{right}})$ \COMMENT{semantic coverage check / repair}
    \IF{$(q_{\text{left}}, q_{\text{right}}) = \bot$}{}
        \STATE \textbf{return} $\bot$ \COMMENT{$\pi(\tilde{x}, s) = \textsc{fail}$: split not repairable}
    \ENDIF

    \STATE $a_{\text{left}} \gets \textsc{Solve}(q_{\text{left}}, \mathcal{H}, d+1)$ \COMMENT{prerequisite branch first; $\mathcal{H}$ updated in place}
    \STATE $\Call{Update}{\mathcal{H}, (q_{\text{left}}, a_{\text{left}})}$ \COMMENT{record the left answer before the dependent branch}
    \STATE $a_{\text{right}} \gets \textsc{Solve}(q_{\text{right}}, \mathcal{H}, d+1)$ \COMMENT{dependent branch grounded by updated $\mathcal{H}$}

    \STATE $a \gets \textsc{Synthesize}(q, a_{\text{left}}, a_{\text{right}}, \mathcal{H}, Q)$
    \STATE \textbf{return} $a$
\ENDFUNCTION
\end{algorithmic}
\end{algorithm}

The procedure terminates, and the depth limit alone bounds its cost.
Execution forms a binary tree rooted at depth $0$ whose maximum depth is $d_{\max}$, so at most $\sum_{j=0}^{d_{\max}} 2^{j} = 2^{d_{\max}+1} - 1$ nodes are visited and the resolution operator runs at most once per visited node.
Expansion, verification, and synthesis occur only at internal nodes, of which there are at most $\sum_{j=0}^{d_{\max}-1} 2^{j} = 2^{d_{\max}} - 1$.
Every recursive call increases the depth by one and no node at depth $d_{\max}$ expands, so every execution path is finite and the recursion stack holds at most $d_{\max} + 1$ frames.
These are worst-case bounds over a fully expanded tree: at $d_{\max} = 4$ they permit $31$ visited nodes, each costing at least one LLM call, against a measured mean of $5.7$ calls per question (Section~\ref{sec:cost}), because most nodes resolve without expanding.

The execution order of the two branches is an invariant of the procedure rather than a property of any particular split.
At every node, \textsc{Solve} appends the current query, its retrieved evidence, and its answer status to $\mathcal{H}$ as soon as the resolution operator returns, before any branching decision is taken; the right child is then invoked only after the left call has returned on that same history.
When the dependent branch begins, $\mathcal{H}$ therefore already holds the current-node record together with every record produced by the prerequisite subtree, including its retrieved evidence and partial context, whether or not that subtree returned an answer.
This makes the dependency constraint of Section~\ref{sec:decomposition} an execution guarantee; it does not assert that the intermediate answers are correct.

\section{Derivation of the STOP/EXPAND Threshold Rule}
\label{sec:threshold_proof}

This section derives the threshold rule of Equation~\ref{eq:threshold} in Section~\ref{sec:granularity_control} and decomposes the excess risk of an imperfect estimator.
A node is \emph{expansion-admissible} when both actions are genuinely available to it, that is, when
\begin{equation*}
x = (q, \mathcal{H}, d) \in \mathcal{X}_{\mathrm{feas}} := \{\, (q, \mathcal{H}, d) \;:\; d < d_{\max} \ \text{and} \ q \ \text{is decomposable} \,\} ;
\end{equation*}
outside $\mathcal{X}_{\mathrm{feas}}$, \textsc{expand} is inadmissible: a node that resolves still stops, and an unresolved one is forced to \textsc{fail}; every statement below is conditioned on admissibility.
Throughout, $\tilde{x} = (x, D)$ is the post-resolution state of an expansion-admissible node, with $D = R_k(q', C)$ the retrieved evidence, and $Z \in \{\mathrm{R}, \mathrm{U}\}$ indicates whether $q$ is resolvable ($\mathrm{R}$) or unresolved ($\mathrm{U}$) from $D$ together with $\mathcal{H}$.
Write $p(\tilde{x}) = \Pr[Z = \mathrm{U} \mid \tilde{x}]$.

\paragraph{Cost model.}
Two actions are available at such a node.
Taking \textsc{expand} at a node that was in fact resolvable incurs $\Delta_{\mathrm{R}}(\tilde{x}) > 0$, and taking \textsc{stop} at a node that was in fact unresolved incurs $\Delta_{\mathrm{U}}(\tilde{x}) > 0$.
Both penalties are cost-to-go quantities measured over the subtree that the action induces, and therefore include descendant retrieval and LLM calls, drift risk, synthesis, and terminal answer loss.
The action that matches the realized $Z$ is taken as the reference and assigned zero excess cost, so the conditional expected costs are
\begin{align*}
\mathbb{E}[\,\mathrm{cost}(\textsc{expand}) \mid \tilde{x}\,] &= \bigl(1 - p(\tilde{x})\bigr)\,\Delta_{\mathrm{R}}(\tilde{x}), \\
\mathbb{E}[\,\mathrm{cost}(\textsc{stop}) \mid \tilde{x}\,] &= p(\tilde{x})\,\Delta_{\mathrm{U}}(\tilde{x}).
\end{align*}
We write $\mathcal{R}(\pi)$ for the expected cost of a policy $\pi$ over the post-resolution states $\tilde{x} = (x, D)$ with $x \in \mathcal{X}_{\mathrm{feas}}$.

\paragraph{Threshold rule.}
\textsc{expand} is optimal exactly when its conditional expected cost is no larger (state arguments omitted):
\begin{equation*}
\bigl(1 - p\bigr)\Delta_{\mathrm{R}} \;\leq\; p\,\Delta_{\mathrm{U}}
\iff
\Delta_{\mathrm{R}} \;\leq\; p\,\bigl(\Delta_{\mathrm{R}} + \Delta_{\mathrm{U}}\bigr)
\iff
p \;\geq\; \theta(\tilde{x}) := \frac{\Delta_{\mathrm{R}}(\tilde{x})}{\Delta_{\mathrm{R}}(\tilde{x}) + \Delta_{\mathrm{U}}(\tilde{x})},
\end{equation*}
which is Equation~\ref{eq:threshold} of Section~\ref{sec:granularity_control}.
The threshold is state-dependent because both penalties are.
Since the penalties are cost-to-go quantities, the resulting decision is node-wise but not myopic; it does not follow, and we do not claim, that greedily applying the rule at every node yields a globally optimal query tree.

\paragraph{Dominance over unconditional routing.}
Before any estimator is fixed, the cost model already separates evidence-conditioned control from policies that commit to one action in advance.
Let $\pi_{\mathrm{orc}}$ be the oracle that observes the realized $Z$ itself, stopping when $Z = \mathrm{R}$ and expanding when $Z = \mathrm{U}$; it is a stronger reference than the $\pi^{*}$ of the next paragraph, which sees only the posterior $p(\tilde{x})$.
Always-stop agrees with $\pi_{\mathrm{orc}}$ except on $\{Z = \mathrm{U}\}$, where it gives up $\Delta_{\mathrm{U}}(\tilde{x})$, and always-expand agrees except on $\{Z = \mathrm{R}\}$, where it gives up $\Delta_{\mathrm{R}}(\tilde{x})$, so
\begin{align*}
\mathcal{R}(\pi_{\mathrm{stop}}) - \mathcal{R}(\pi_{\mathrm{orc}}) &= \Pr[Z = \mathrm{U}] \, \mathbb{E}\bigl[\Delta_{\mathrm{U}}(\tilde{x}) \mid Z = \mathrm{U}\bigr], \\
\mathcal{R}(\pi_{\mathrm{exp}}) - \mathcal{R}(\pi_{\mathrm{orc}}) &= \Pr[Z = \mathrm{R}] \, \mathbb{E}\bigl[\Delta_{\mathrm{R}}(\tilde{x}) \mid Z = \mathrm{R}\bigr] .
\end{align*}
Each difference is strictly positive whenever its state occurs with positive probability.
This is the value of observing the evidence state, realized by the oracle; an operational policy realizes it only to the extent that the excess risk of its estimator, derived next, stays below these margins.
Appendix~\ref{sec:query_routing} measures the converse: a learned router that never observes the evidence collapses onto a near-constant policy and scores below that constant policy itself.

\paragraph{Excess risk of an estimator.}
Let $\hat{Z}$ be any estimator inducing a policy $\pi_{\hat{Z}}$, and let $\pi^{*}$ be the rule above.
The two actions differ in conditional expected cost by
\begin{equation*}
\bigl(1 - p\bigr)\Delta_{\mathrm{R}} - p\,\Delta_{\mathrm{U}}
= \bigl(\Delta_{\mathrm{R}} + \Delta_{\mathrm{U}}\bigr)\bigl(\theta - p\bigr),
\end{equation*}
so disagreeing with $\pi^{*}$ at $\tilde{x}$ costs $(\Delta_{\mathrm{R}} + \Delta_{\mathrm{U}})\,\lvert p - \theta \rvert$ and agreeing costs nothing.
Taking expectations,
\begin{equation*}
\mathcal{R}(\pi_{\hat{Z}}) - \mathcal{R}(\pi^{*})
= \mathbb{E}\Bigl[\bigl(\Delta_{\mathrm{R}}(\tilde{x}) + \Delta_{\mathrm{U}}(\tilde{x})\bigr)\,\bigl\lvert p(\tilde{x}) - \theta(\tilde{x}) \bigr\rvert \cdot \mathbf{1}\{\pi_{\hat{Z}}(\tilde{x}) \neq \pi^{*}(\tilde{x})\}\Bigr].
\end{equation*}
The indicator splits into the two error directions: \emph{premature stopping}, where $p > \theta$ but the estimator stops, and \emph{unnecessary expansion}, where $p < \theta$ but the estimator expands.
Each contributes in proportion to how far $p$ lies from the threshold, so errors on states where the evidence is genuinely ambiguous are cheap and errors on clear-cut states are expensive.
This is the sense in which the resolution test is an estimator of a defined decision rather than a heuristic: it is a hard classifier $\hat{Z}$ whose excess risk is governed by the expression above, and any calibrated or trained replacement is evaluated on the same scale.

\paragraph{Why the estimator must observe the evidence state.}
The running example of Figure~\ref{fig:overview} instantiates the difficulty.
For a fixed corpus and a deterministic retriever, a fixed query yields a single evidence set, so the argument is made over the distribution of problem instances $(Q, C)$ on which a policy is evaluated: the same question can appear against different corpora, or different versions of a corpus, that rank its evidence differently.
Take $q^{*}$ to be ``When was the start of the battle of the birthplace of the performer of III?'', and hold the retriever and the gold answer fixed.
In an instance whose root top-$k$ contains the album, performer, birthplace, and battle passages, the node is resolvable and \textsc{stop} is optimal.
In an instance whose top-$k$ omits the passage naming the performer, which is still in the corpus and merely not ranked into the budget, \textsc{expand} is optimal, because the dependent sub-queries retrieve against a narrower target.
The query is identical in the two instances and the difference lies entirely in which passages the retrieval budget surfaced, so no function of $q^{*}$ alone separates them.
Consider two post-resolution states $\tilde{x}_1 = (x, D_1)$ and $\tilde{x}_2 = (x, D_2)$ that share the search state $x$ but differ in the retrieved evidence, $D_1 \neq D_2$, and suppose $\pi^{*}(\tilde{x}_1) = \textsc{stop}$ while $\pi^{*}(\tilde{x}_2) = \textsc{expand}$, each occurring with positive probability under the instance distribution.
Any policy measurable with respect to $q$ alone assigns the same action to both, and therefore disagrees with $\pi^{*}$ on at least one of them.
Its excess risk is bounded below by
\begin{equation*}
\min_{i \in \{1,2\}} \Pr[\tilde{x}_i] \cdot \bigl(\Delta_{\mathrm{R}}(\tilde{x}_i) + \Delta_{\mathrm{U}}(\tilde{x}_i)\bigr) \bigl\lvert p(\tilde{x}_i) - \theta(\tilde{x}_i) \bigr\rvert \;>\; 0 ,
\end{equation*}
a quantity that no amount of query-only training data reduces, since the two states are indistinguishable to such a policy.
Learning is therefore not what separates \Ours from query-level routing; conditioning on the realized evidence state is.
Appendix~\ref{sec:learned_router} reports what happens when a learned, calibrated estimator that does observe $\tilde{x}$ replaces the resolution test.

\section{A Learned and Calibrated STOP/EXPAND Policy}
\label{sec:learned_router}

\Ours's routing decision is implemented as a training-free test: a node is expanded when the resolution operator returns \textsc{unresolved}.
This section reports a controlled experiment showing that the same decision admits a learned, calibrated estimator, and that replacing the training-free test with one changes neither accuracy nor cost enough to justify the added machinery.
The experiment therefore establishes that the STOP/EXPAND interface is instantiable by a trained controller, not that a trained controller is required.

All runs use the MuSiQue controlled-corpus setting.
For each eligible saved state we force both STOP and EXPAND through to a final answer and label the state with the lower-loss action, collecting $3{,}060$ labelled decision states from $817$ training questions.
The LLM, the retriever, and the NV-Embed-v2 encoder are frozen, and only a logistic action head is trained on the resulting state representations.
Model selection, Platt calibration, and testing use disjoint sets of $200$, $400$, and $1{,}000$ questions; the $1{,}417$ questions used for training, selection, and calibration are drawn from the MuSiQue development pool outside the evaluation subset.

On held-out decision states, Platt calibration reduces expected calibration error from $0.111$ to $0.082$ and decision regret by $0.024$ (95\% CI $[0.005, 0.047]$).
The action head is therefore not merely a classifier of convenience: its probabilities carry usable calibration, which is what the threshold form of the routing decision requires.

We then replace only the root STOP/EXPAND rule, leaving every other component at the configuration used in the paper, and rerun all $1{,}000$ test questions.
Relative to the training-free policy, two thresholds on the calibrated probability, one chosen for accuracy alone and one that also weights cost, lose $0.30$ EM (95\% CI $[-2.10, +1.50]$) and $1.40$ EM (95\% CI $[-3.20, +0.40]$) respectively; neither difference is statistically significant.
LLM calls per question rise from $7.63$ to $10.41$ for the accuracy-only threshold and to $8.64$ for the cost-weighted one.
The learned estimator thus neither improves accuracy nor lowers cost, so we retain the training-free configuration.
This is a root-level proof of instantiation rather than a learned replacement for every recursive decision, and we do not claim that the trained head computes the posterior of the underlying decision.

\section{Query-Level Routing}
\label{sec:query_routing}

Adaptive-RAG~\cite{adaptiverag} predicts question complexity before retrieval and selects a global no-, single-, or multi-step strategy.
We train its classifier on the $1{,}417$ MuSiQue development questions outside our evaluation subset, so the router operates in-distribution, and evaluate it on MuSiQue-full with a shared Qwen3.6-27B reader and retrieval stack; the numbers are therefore not comparable with Table~\ref{tab:full_corpus}.
It falls far short of \Ours (Table~\ref{tab:query_routing}).
The routing behavior is more informative than the gap: $959$ of the $1{,}000$ questions are routed to multi-step retrieval, $41$ to single-step, and none to no-retrieval, and a control policy that always selects multi-step scores above Adaptive-RAG itself.
The learned routing decision therefore contributes nothing over a constant policy.
Whether a query is resolvable at its current granularity depends on the evidence actually returned, not on the apparent complexity of the original question.
A controller that conditions on that evidence remains compatible with \Ours, as Appendix~\ref{sec:learned_router} shows.

\begin{table}[h]
\centering
\caption{Query-level routing on MuSiQue-full, under a shared Qwen3.6-27B reader and retrieval stack.
\emph{Always multi-step} is a constant-policy control for Adaptive-RAG.
These values are not comparable with Table~\ref{tab:full_corpus}, which uses GPT-4o-mini.
Bold indicates the best value in each column.}
\label{tab:query_routing}
\footnotesize
\setlength{\tabcolsep}{8pt}
\renewcommand{\arraystretch}{0.95}
\begin{tabular}{lcc}
\toprule
Method & EM & F1 \\
\midrule
Adaptive-RAG~\cite{adaptiverag}          & 26.4 & 33.9 \\
\quad \emph{Always multi-step} (control) & 27.3 & 34.6 \\
\midrule
\textbf{\Ours}                           & \textbf{47.9} & \textbf{59.3} \\
\bottomrule
\end{tabular}
\end{table}

\section{Evaluation Details}
\label{sec:eval_details}

Table~\ref{tab:full_corpus} marks reproduced baselines with an asterisk, but the asterisk alone does not make the decision rule auditable.
We quote a published number only when the source reports the same dataset subset, corpus regime, reader setting, retriever, and metric protocol as our evaluation, and otherwise reproduce the method ourselves.
The criterion is evaluation compatibility, not the age or expected strength of a baseline.

Every controlled-corpus row of Table~\ref{tab:full_corpus}, quoted or reproduced, uses the same 1{,}000-question subsets per dataset, GPT-4o-mini as the reader, NV-Embed-v2 as the retriever with $k=5$, and the pooled Recall@$k$ protocol defined below.
Retrieval for every run is performed on one NVIDIA Tesla P40 GPU, except HotpotQA-full, which uses four NVIDIA RTX A6000 GPUs; reading is served by the OpenAI API.
Retrieval scores passages by L2-normalized dot product, and the depth limit $d_{\max}=4$ matches the maximum reasoning depth analyzed in MuSiQue and bounds recursive cost.

Recall@$k$ for $k \in \{2,5\}$ measures whether the gold supporting documents appear in the retrieved top-$k$ passages.
For multi-query methods, we pool the passages retrieved across all node- or step-level queries, deduplicate them by retaining the highest embedding score, and globally re-rank the pool before computing Recall@$k$; this keeps the evaluation budget $k$ fixed across methods while making retrieval volume explicit through the call and token statistics in Table~\ref{tab:cost_analysis}.
Because all runs use temperature-zero decoding, the bootstrap intervals reported in Section~\ref{sec:experiments} quantify uncertainty over the question population rather than run-to-run variability.

Cost accounting for Table~\ref{tab:cost_analysis}: API cost is priced at GPT-4o-mini list prices as of August 2026, \$0.15 per 1M input tokens and \$0.60 per 1M output tokens, which is the basis of the $8.5\times$ figure in Section~\ref{sec:cost}.
The cost-matched configuration caps recursion at depth $1$, so the root may still be expanded once but its children are not.
Its EM gain over IRCoT holds on every benchmark, from $+3.5$ to $+18.1$, and its output generation nonetheless remains $77\%$ higher than IRCoT's ($131$ versus $74$ tokens), so that operating point is cheaper without being uniformly lighter.

The quoted entries are taken from the GPT-4o-mini rows of \cite{hipporag2}, which uses the same subsets and retriever, with the QA numbers and the passage recall coming from two different appendix tables of that paper.
GraphRAG has no recall entry there, since it does not directly produce passage retrieval results, which is why its recall cells are empty in Table~\ref{tab:full_corpus}.
All full-corpus results in Table~\ref{tab:full_corpus} are reproduced, as no prior work reports these methods on the complete corpora under our protocol.
The coding agent and the Recursive Language Model are run under the same standardization, which confines the comparison to their control logic; because the original studies use stronger models and different native corpus interfaces, these are controlled adaptations rather than reproductions of their published configurations.

\begin{table}[h]
\centering
\caption{Source of every baseline number in Table~\ref{tab:full_corpus}.
All rows share the reader, retriever, subsets, and metric protocol described above.}
\label{tab:baseline_provenance}
\footnotesize
\setlength{\tabcolsep}{5pt}
\renewcommand{\arraystretch}{0.95}
\begin{tabular}{llll}
\toprule
Method & Corpus regime & EM / F1 & Recall@2 / @5 \\
\midrule
NV-Embed-v2~\cite{nvembed}  & controlled                & Quoted & Quoted \\
RAPTOR~\cite{raptor}        & controlled                & Quoted & Quoted \\
GraphRAG~\cite{graphrag}    & controlled                & Quoted & Not reported \\
HippoRAG~\cite{hipporag}    & controlled                & Quoted & Quoted \\
HippoRAG 2~\cite{hipporag2} & controlled                & Quoted & Quoted \\
\midrule
PropRAG~\cite{proprag}      & controlled, MuSiQue-full, 2Wiki-full & Reproduced & Reproduced \\
Self-Ask~\cite{selfask}     & controlled, full          & Reproduced & Reproduced \\
Least-to-Most~\cite{leasttomost} & controlled, full     & Reproduced & Reproduced \\
IRCoT~\cite{ircot}          & controlled, full          & Reproduced & Reproduced \\
Coding agent~\cite{codingagent} & full                  & Reproduced & Reproduced \\
RLM~\cite{rlm}              & full                      & Reproduced & Reproduced \\
\midrule
\Ours                       & controlled, full          & This work  & This work \\
\bottomrule
\end{tabular}
\end{table}

\section{Robustness across Reader Backbones}
\label{sec:open_source_results}
\Ours's gains are not specific to a single reader backbone.
To test reader robustness, we re-run the evaluation pipeline in the controlled-corpus setting with two additional LLM readers spanning different architectures: Llama-3.3-70B~\cite{llama3} and Qwen3-30B-A3B~\cite{qwen3}. Tables~\ref{tab:qa_performance_llm} and~\ref{tab:qa_performance_qwen3} show that \Ours's gains are not specific to a single reader backbone.
Across two additional readers, Llama-3.3-70B and Qwen3-30B-A3B, \Ours retains the best average QA performance, achieving $60.2$ EM / $71.2$ F1 and $58.3$ EM / $68.9$ F1, respectively.
It outperforms the strongest reproduced graph-based baseline, PropRAG, by $+3.6$ EM with Llama-3.3-70B and $+11.0$ EM with Qwen3-30B-A3B, while also outperforming iterative/decomposition baselines.
This supports that evidence-conditioned refinement is robust across dense and MoE reader architectures.

\begin{table*}[ht]
\caption{QA performance with Llama-3.3-70B~\cite{llama3} in the controlled-corpus setting.
Bold indicates the best value in each column.
$^{*}$Results are reproduced.}
\label{tab:qa_performance_llm}
\centering
\footnotesize
\renewcommand{\arraystretch}{0.95}
\setlength{\tabcolsep}{5pt}
\begin{tabular}{l|cc|cc|cc|cc}
\toprule
\multirow{2}{*}{Method}
& \multicolumn{2}{c|}{HotpotQA}
& \multicolumn{2}{c|}{MuSiQue}
& \multicolumn{2}{c|}{2Wiki}
& \multicolumn{2}{c}{Avg}\\

\cmidrule(lr){2-3} \cmidrule(lr){4-5} \cmidrule(lr){6-7} \cmidrule(lr){8-9}

& EM & F1 & EM & F1 & EM & F1 & EM & F1 \\
\midrule
NV-Embed-v2          & 62.8 & 75.3 & 34.7 & 45.7 & 57.5 & 61.5 & 51.7 & 60.8 \\
\midrule
RAPTOR               & 56.8 & 69.5 & 20.7 & 28.9 & 47.3 & 52.1 & 41.6 & 50.2 \\
GraphRAG             & 55.2 & 68.6 & 27.3 & 38.5 & 51.4 & 58.6 & 44.6 & 55.2 \\
HippoRAG             & 52.6 & 63.5 & 26.2 & 35.1 & 65.0 & 71.8 & 47.9 & 56.8 \\
HippoRAG 2           & 62.7 & 75.5 & 37.2 & 48.6 & 65.0 & 71.0 & 55.0 & 65.0 \\
PropRAG$^{*}$        & 62.6 & 76.0 & 41.5 & 53.6 & 65.6 & 74.3 & 56.6 & 68.0 \\
\midrule
Self-Ask$^{*}$       & 58.8 & 71.7 & 33.9 & 47.8 & 64.9 & 74.9 & 52.5 & 64.8 \\
Least-to-Most$^{*}$  & 57.6 & 72.9 & 29.4 & 43.5 & 44.2 & 48.9 & 43.7 & 55.1 \\
IRCoT$^{*}$          & 59.7 & 72.0 & 31.2 & 40.7 & 68.6 & 78.3 & 53.2 & 63.7 \\
\midrule
\textbf{\Ours}       & \textbf{65.4} & \textbf{78.0} & \textbf{45.0} & \textbf{56.6} & \textbf{70.3} & \textbf{78.9} & \textbf{60.2} & \textbf{71.2} \\
\bottomrule
\end{tabular}
\end{table*}

\begin{table*}[ht]
\caption{QA performance with Qwen3-30B-A3B~\cite{qwen3} in the controlled-corpus setting.
Bold indicates the best value in each column.
$^{*}$Results are reproduced.}
\label{tab:qa_performance_qwen3}
\centering
\footnotesize
\renewcommand{\arraystretch}{0.95}
\setlength{\tabcolsep}{5pt}
\begin{tabular}{l|cc|cc|cc|cc}
\toprule
\multirow{2}{*}{Method}
& \multicolumn{2}{c|}{HotpotQA}
& \multicolumn{2}{c|}{MuSiQue}
& \multicolumn{2}{c|}{2Wiki}
& \multicolumn{2}{c}{Avg}\\
\cmidrule(lr){2-3} \cmidrule(lr){4-5} \cmidrule(lr){6-7} \cmidrule(lr){8-9}
& EM & F1 & EM & F1 & EM & F1 & EM & F1 \\
\midrule
PropRAG$^{*}$        & 55.0 & 69.2 & 32.2 & 44.7 & 54.6 & 63.8 & 47.3 & 59.2 \\
\midrule
Self-Ask$^{*}$       & 52.8 & 67.3 & 37.6 & 47.9 & 66.0 & \textbf{75.2} & 52.1 & 63.5 \\
Least-to-Most$^{*}$  & 53.8 & 67.9 & 33.1 & 43.1 & 37.4 & 42.5 & 41.4 & 51.2 \\
IRCoT$^{*}$          & 62.9 & 75.3 & 34.4 & 44.8 & 52.6 & 58.5 & 50.0 & 59.5 \\
\midrule
\textbf{\Ours}       & \textbf{65.1} & \textbf{77.3} & \textbf{43.4} & \textbf{54.6} & \textbf{66.3} & 74.8 & \textbf{58.3} & \textbf{68.9} \\
\bottomrule
\end{tabular}
\end{table*}

\section{Robustness to Embedding Models}
\label{sec:embedding_robustness}

\Ours's retrieval gains are not tied to a single embedding backbone.
Table~\ref{tab:embedding_comparison_full} evaluates \Ours with text-embedding-3-large, Qwen3-Embedding-8B~\cite{qwen3}, and NV-Embed-v2.
Across the three retrievers, \Ours improves average Recall@5 by $9.4$, $10.3$, and $8.2$ points, respectively, over dense retrieval alone.
NV-Embed-v2 gives the strongest overall recall, which motivates its use as the default encoder.
The consistent gains across backbones indicate that \Ours's improvement comes from evidence-guided query refinement rather than from a particular embedding model.

\begin{table}[h]
\centering
\caption{Average Recall@5 across embedding models.
Bold indicates the best value in each column.}
\label{tab:embedding_comparison_full}
\footnotesize
\setlength{\tabcolsep}{8pt}
\renewcommand{\arraystretch}{0.95}
\begin{tabular}{lcc}
\toprule
Retriever & Dense & \Ours \\
\midrule
text-embedding-3-large & 74.5 & 83.9 \\
Qwen3-Embedding-8B     & 68.7 & 79.0 \\
NV-Embed-v2            & \textbf{80.2} & \textbf{88.4} \\
\bottomrule
\end{tabular}
\end{table}

\section{Trigger Rate and Round-trip Repair Statistics}
\label{sec:trigger_rate}

Trigger and repair statistics show that \Ours decomposes mainly on benchmarks where query--evidence granularity mismatch is more frequent, and that verification is actively used rather than acting as a passive check.
Table~\ref{tab:trigger-rate} reports how often decomposition is triggered and how often round-trip verification repairs a proposed decomposition across datasets.
The trigger rate is much lower on HotpotQA ($10.7\%$) than on MuSiQue ($45.2\%$) and 2Wiki ($48.6\%$), which is consistent with HotpotQA's greater shortcut availability and relatively simpler reasoning structure.
Round-trip repair occurs in $14.8$--$18.8\%$ of triggered cases, showing that the verifier corrects imperfect decompositions in a non-trivial fraction of recursive calls.

\begin{table}[h]
\centering
\caption{Decomposition trigger rate and round-trip repair frequency across datasets.}
\label{tab:trigger-rate}
\footnotesize
\setlength{\tabcolsep}{8pt}
\begin{tabular}{lcc}
\toprule
Dataset & Trigger (\%) & Repair (\%) \\
\midrule
HotpotQA & 10.7 & 18.8 \\
MuSiQue & 45.2 & 14.8 \\
2Wiki & 48.6 & 17.0 \\
\bottomrule
\end{tabular}
\end{table}

\section{Strict Evidence-grounded Evaluation}
\label{sec:strict_eval}

\Ours's gains are not an artifact of parametric memorization by the reader.
Although \Ours encourages evidence-grounded reasoning, the default evaluation does not require every intermediate answer to cite annotated gold documents.
To test whether this permissiveness inflates performance, we evaluate a stricter variant that requires document-level grounding for each answer on a filtered subset of 634 MuSiQue queries, where gold annotations are reliable enough for strict citation enforcement~\cite{selfrag, mainrag}.
As shown in Table~\ref{tab:evidence-grounded}, the strict variant achieves nearly identical performance to \Ours.
Moreover, all baselines in Table~\ref{tab:full_corpus} use the same GPT-4o-mini reader and likewise enforce no document-level citation or entailment constraint during answering, so any contribution from parametric knowledge should affect all methods similarly.
The differential gains of \Ours over these baselines therefore reflect evidence-conditioned refinement and decomposition control rather than parametric memorization.

\begin{table}[h]
\centering
\caption{Evaluation of a stricter evidence-grounded variant on the MuSiQue refined subset.
Bold indicates the best value in each column.}
\label{tab:evidence-grounded}
\footnotesize
\renewcommand{\arraystretch}{0.95}
\setlength{\tabcolsep}{6pt}
\begin{tabular}{lcc}
\toprule
Setting & EM & F1 \\
\midrule
\Ours & 47.5 & \textbf{61.9} \\
Strict evidence-grounded variant & \textbf{47.6} & 61.7 \\
\bottomrule
\end{tabular}
\end{table}

\paragraph{Closed-book control.}
As an additional control for parametric knowledge, we evaluate GPT-4o-mini without retrieval, providing only the question as input on the same datasets.
As shown in Table~\ref{tab:closed_book}, the parametric-only model performs substantially below \Ours across all datasets, with 13.1 EM / 21.7 F1 on MuSiQue, 23.1 EM / 28.8 F1 on 2Wiki, and 29.4 EM / 39.7 F1 on HotpotQA.
Although the model can answer some questions from memorized knowledge, these scores are far below the retrieval-augmented results, especially on MuSiQue where genuine multi-hop reasoning is required.
Together with the strict grounding variant and the shared-reader comparison, this supports that \Ours's gains are driven primarily by evidence acquisition and decomposition control rather than parametric memorization.

\begin{table}[h]
\centering
\caption{Closed-book GPT-4o-mini performance on the 1{,}000-question subsets.}
\label{tab:closed_book}
\footnotesize
\renewcommand{\arraystretch}{0.95}
\setlength{\tabcolsep}{6pt}
\begin{tabular}{lcc}
\toprule
Dataset & EM & F1 \\
\midrule
MuSiQue & 13.1 & 21.7 \\
2Wiki & 23.1 & 28.8 \\
HotpotQA & 29.4 & 39.7 \\
\bottomrule
\end{tabular}
\end{table}

\paragraph{Backbone sensitivity.}
The controls above hold the reader fixed at GPT-4o-mini, and that choice is deliberate.
Repeating the closed-book control on MuSiQue with GPT-5 yields $25.0$ EM / $37.6$ F1, well above GPT-4o-mini's $13.1$ / $21.7$, and when GPT-5 refines sub-queries it frequently supplies bridge entities that were never retrieved.
A stronger backbone therefore answers a larger share of questions from parametric knowledge, which makes answer accuracy a weaker diagnostic of retrieval--query alignment.
Holding the reader at GPT-4o-mini keeps model capability fixed and attributes differences between methods to their control and decomposition logic instead.
We report this as a sensitivity analysis of the backbone choice, not as a performance ceiling for \Ours.

\section{Comparison with Dedicated Decomposition Baselines}
\label{sec:decomp_comparison}
This comparison tests whether \Ours remains competitive with dedicated decomposition methods under their reported evaluation setting.
We focus on TRQA~\cite{trqa} because it is the most recent state-of-the-art query-decomposition baseline among the methods discussed in Section~\ref{sec:related_work}.
Since both Q-DREAM~\cite{qdream} and TRQA are fine-tuned methods, we select TRQA as the stronger fine-tuned comparison.
We also include Self-Ask~\cite{selfask} and Least-to-Most~\cite{leasttomost} as prominent prompt-based decomposition baselines.
Decomposed Prompting~\cite{decomposed_prompting} is excluded because its number of LLM calls is computationally prohibitive in our evaluation setting.
Because TRQA is closed-source, we match its reported setting---GPT-3.5-Turbo reader, ColBERTv2 retriever, the full MuSiQue development set of $2{,}417$ questions over the complete corpus, and the Answer Recall metric---and cite its reported score for a standardized comparison; \Ours is therefore run on all $2{,}417$ questions here rather than on the $1{,}000$-question subset used elsewhere.
Under this setting, \Ours reaches $29.3$ Answer Recall, improving over TRQA's $26.8$ and over the prompt-based baselines Least-to-Most and Self-Ask, which obtain $13.4$ and $15.6$, respectively (Table~\ref{tab:decomp_comparison}).

\begin{table}[!h]
\caption{Comparison against dedicated decomposition baselines on MuSiQue-full ($N{=}2{,}417$) under TRQA's reported setting (GPT-3.5-Turbo reader, ColBERTv2 retriever). The metric is \emph{Answer Recall} as defined and reported by TRQA; the baseline scores are those reported by TRQA, and \Ours is measured in this work on the same questions with the same reader and retriever. Bold indicates the best value.}
\label{tab:decomp_comparison}
\centering
\footnotesize
\setlength{\tabcolsep}{8pt}
\begin{tabular}{lc}
\toprule
Method & Answer Recall \\
\midrule
Least-to-Most~\cite{leasttomost} (ICLR 2023) & 13.4 \\
Self-Ask~\cite{selfask} (EMNLP 2023) & 15.6 \\
TRQA~\cite{trqa} (AAAI 2024) & 26.8 \\
\textbf{\Ours} & \textbf{29.3} \\
\bottomrule
\end{tabular}
\end{table}

\section{Voting over Multiple Decompositions}
\label{sec:multi_decomp}

\Ours commits to a single decomposition at each expansion, which raises the question of whether an early root-level split that is locally plausible but poorly aligned with the corpus propagates to the final answer.
This section tests the natural remedy: generating several root decompositions and letting them vote.
On MuSiQue-full ($1{,}000$ questions, GPT-4o-mini, the same retrieval stack as the rest of the paper), we generate $b$ alternative root-level decompositions with distinct pivots, run each candidate independently through the full pipeline, and select the final answer by majority vote with a judge tie-break.
Here $b=1$ is the configuration used throughout the paper and reproduces its MuSiQue-full result in Table~\ref{tab:full_corpus}.

Accuracy increases monotonically but modestly, while token cost grows close to linearly in $b$ (Table~\ref{tab:multi_decomp}).
Wall-clock latency need not grow proportionally, since the $b$ candidates are independent and can be executed in parallel.
That the gains are small is itself informative: it indicates that early decomposition errors rarely survive to the final answer.
A sub-query that cannot be answered from the corpus returns unresolved rather than committing a wrong intermediate value, so a poor split tends to be absorbed by further refinement or by the verifier instead of being propagated.
Multiple root decompositions are therefore a valid extension of \Ours rather than a correction to it, and $b=1$ remains the better cost-effectiveness point, which is why we retain it.

\begin{table}[h]
\centering
\caption{Root-level decomposition candidates on MuSiQue-full.
Cost is the per-question API token cost relative to $b=1$, the configuration used throughout the paper.}
\label{tab:multi_decomp}
\footnotesize
\setlength{\tabcolsep}{8pt}
\renewcommand{\arraystretch}{0.95}
\begin{tabular}{cccc}
\toprule
$b$ & EM & F1 & Cost \\
\midrule
1 & 37.4 & 50.6 & $1.0\times$ \\
2 & 38.3 & 51.4 & $1.8\times$ \\
3 & 39.0 & 52.2 & $2.6\times$ \\
\bottomrule
\end{tabular}
\end{table}

\section{Binary Decomposition under High Arity}
\label{sec:high_arity}

Binary expansion is an execution primitive, not an assumption that every question has a single linear reasoning chain.
A plan with $m$ leaf information needs is organized as at most $m-1$ binary reductions, with the expansion operator, the verifier, and the execution framework unchanged; how many reductions actually occur is decided by the controller, since a node resolvable from its retrieved passages terminates there.
This section tests whether that primitive degrades as the number of leaf needs grows.

No question among our $1{,}000$ MuSiQue questions has a gold reasoning node with immediate fan-in $3$ or higher, so we construct $400$ controlled questions from $100$ five-fact MuSiQue bundles with nested variants for $m = 2, 3, 4, 5$.
Each branch carries one gold passage and three mined distractors, holding gold density at $25\%$ so that retrieval difficulty stays constant as arity grows.
The set was frozen before evaluation, and two annotators verified $100$ sampled questions and their answers.
Forced expansion prevents the controller from terminating without decomposing, so the expansion operator itself is under test rather than the routing decision; the oracle-plan control supplies gold leaf queries while retrieval and reading still run.

Branch generation is not the bottleneck.
Table~\ref{tab:high_arity} shows that binary \Ours stays between $95.0$ and $99.0$ EM as arity rises from $2$ to $5$, losing $0.80$ EM per additional branch, and its generated decompositions are statistically indistinguishable from the binary oracle-plan control ($-1.50$ EM, 95\% CI $[-3.50, +0.50]$).
Variable-arity branching gives no significant mean-EM improvement over binary decomposition (binary $-$ variable $= +0.25$, 95\% CI $[-2.00, +2.50]$), although it does reduce latency.
IRCoT degrades fastest, at $5.00$ EM per additional branch.

Arity is costly in aggregation rather than in branching.
A separate comparison varies only the final combination step: both arms receive the gold plan and run identical retrieval and per-branch reading, but one combines by sequential binary reduction and the other by a single flat $N$-ary aggregation.
The flat variant loses $27.50$ EM ($71.0$ versus $98.5$ overall; 95\% CI $[24.00, 31.00]$), so the measured trade-off of binary refinement is additional sequential cost, not an observed high-arity accuracy failure, and we retain dependency-preserving binary decomposition.

Ordinary comparison questions are covered by the same primitive.
The class often cited as requiring parallel aggregation, retrieving an attribute for one entity, retrieving the corresponding attribute for another, and comparing them, is represented by two branches followed by synthesis, with each resolved value written to $\mathcal{H}$ before the dependent branch runs.
On the $235$ comparison questions in 2Wiki, \Ours reaches $85.5$ EM, which together with the depth-stratified results in Section~\ref{sec:experiments} indicates that ordinary two-source comparison is within the evaluated scope.

\begin{table}[h]
\vspace{-2mm}
\centering
\caption{Forced-expansion results on the $400$-question controlled high-arity set, where $m$ is the number of leaf information needs.
EM slope is the mean EM change per additional branch.
In the binary conditions every \textsc{expand} emits exactly two dependency-ordered children; the variable-arity condition lets it emit up to $m$ children.}
\label{tab:high_arity}
\footnotesize
\setlength{\tabcolsep}{6pt}
\renewcommand{\arraystretch}{0.95}
\begin{tabular}{lccccc}
\toprule
Method & EM@2 & EM@3 & EM@4 & EM@5 & EM slope \\
\midrule
IRCoT                            & 100.0 & 99.0 & 94.0 & 85.0 & $-5.00$ \\
Oracle-plan, binary              & 100.0 & 99.0 & 98.0 & \textbf{97.0} & $-1.00$ \\
\Ours, variable-arity (forced)   & 100.0 & 98.0 & \textbf{99.0} & 90.0 & $-2.90$ \\
\Ours, binary (forced)           & 97.0  & \textbf{99.0} & 97.0 & 95.0 & $\mathbf{-0.80}$ \\
\bottomrule
\end{tabular}
\vspace{-4mm}
\end{table}

\section{Comparison under a Code-Capable Backbone}
\label{sec:agentic_comparison}

PyRAG~\cite{pyrag} specializes the corpus-as-environment paradigm for multi-hop RAG, representing reasoning as an executable program over retrieval and answering tools, which makes intermediate state explicit and supports execution-grounded repair.
It requires a code-capable backbone, so we evaluate it and \Ours with a shared Qwen3.6-27B reader and retrieval stack on MuSiQue-full, a setting not comparable with Table~\ref{tab:full_corpus}: \Ours reaches $47.9$ EM / $59.3$ F1 against PyRAG's $41.2$ / $52.9$ (Table~\ref{tab:agentic_comparison}).
Giving the agentic baseline a stronger code model therefore does not change the ordering.

\begin{table}[h]
\centering
\caption{Comparison with PyRAG on MuSiQue-full, under a shared Qwen3.6-27B reader and retrieval stack.
These values are not comparable with Table~\ref{tab:full_corpus}, which uses GPT-4o-mini.
Bold indicates the best value in each column.}
\label{tab:agentic_comparison}
\footnotesize
\setlength{\tabcolsep}{8pt}
\renewcommand{\arraystretch}{0.95}
\begin{tabular}{lcc}
\toprule
Method & EM & F1 \\
\midrule
PyRAG~\cite{pyrag}  & 41.2 & 52.9 \\
\midrule
\textbf{\Ours}      & \textbf{47.9} & \textbf{59.3} \\
\bottomrule
\end{tabular}
\end{table}

\section{Limitations}
\label{sec:limitations}
\vspace{-2mm}
Our experiments establish performance on English, Wikipedia-derived, passage-based short-answer multi-hop QA, and the results should not be read as evidence of transfer beyond that setting.
The dependency structures the evaluation covers are acyclic and resolvable branch by branch; questions whose prerequisites are mutually dependent, or whose branches must be optimized jointly rather than settled in sequence, are outside what the current state representation models.
The untested dimensions have established benchmarks of their own: long-form QA (ASQA~\cite{asqa}, ELI5~\cite{eli5}), multilingual QA (XOR-TyDi QA~\cite{xorqa}, MKQA~\cite{mkqa}), domain-specific QA (BioASQ~\cite{bioasq}, FinanceBench~\cite{financebench}), structured table--text QA (HybridQA~\cite{hybridqa}, OTT-QA~\cite{ottqa}), and multimodal QA (MultiModalQA~\cite{mmqa}, WebQA~\cite{webqa}).
None is a drop-in extension: the long-form, multilingual, and domain-specific settings change the answer format, language-access assumptions, or corpus domain without controlling for the passage dependency structure studied here, and the structured and multimodal ones additionally require heterogeneous evidence representations and retrieval operators.

\Ours depends on the reliability of LLM modules for resolution, decomposition, verification, and synthesis.
Reader errors can affect the refinement policy: false abstention may cause over-refinement, while unsupported answer generation may stop refinement too early.
Although our calibration and robustness analyses suggest these errors are bounded in the evaluated setting, the current binary resolved/unresolved signal does not capture finer states such as partial evidence, conflicting evidence, or multiple competing bridge entities.
Appendix~\ref{sec:learned_router} replaces that test with a calibrated classifier without changing what the signal can express; representing partial or conflicting evidence would instead require a richer state, such as entailment-based checks over the retrieved passages.

\Ours expands a node only when the reader cannot answer from the retrieved evidence, so on queries the corpus cannot support it abstains rather than fabricates (49 of 50 adversarially unanswerable queries, Figure~\ref{fig:trigger-analysis}(b)).
The same dependence on evidence is the main risk: the verifier checks that a split preserves the question, not that the retrieved passages are true, so a biased or inaccurate corpus yields answers that are well supported and wrong in the same way.
Deployments should therefore use curated corpora and expose the retrieved passages alongside each answer.

\section{Prompt Templates}
\label{sec:Prompts}

This appendix provides a consolidated overview of the prompt templates used to instantiate the operators in Algorithm~\ref{alg:ours}.
Rather than embedding the full prompt specifications in the main text, we summarize their roles here---grouped by the operator they realize---and present the concrete templates in the corresponding figures.

\paragraph{Resolution operator $\mathcal{G}$.}
The resolution operator is realized by three prompts that jointly perform query refinement, retrieval-grounded answering, and resolution-status determination:
\begin{itemize}
    \item \textbf{Subquery Refinement}, which rewrites a query using the accumulated interaction history $\mathcal{H}$ so that implicit references and abstract expressions are grounded in previously resolved facts before retrieval (Figure~\ref{fig:subquery_refinement_prompt}).

    \item \textbf{Granularity Assessment (Root Version)}, which evaluates whether the original query $Q$ can be answered directly under the retrieved evidence and returns either an answer or an unresolved-support signal (Figure~\ref{fig:granularity_assessment_root_prompt}).

    \item \textbf{Granularity Assessment (Recursive Version)}, which applies the same evidence-conditioned resolution check to each internal sub-query node during recursion (Figure~\ref{fig:granularity_assessment_recursive_prompt}).
\end{itemize}

\paragraph{Binary expansion operator $\mathcal{B}$.}
\begin{itemize}
    \item \textbf{Binary Query Decomposition}, which proposes a dependency-ordered pair of sub-queries $(q_{\text{left}}, q_{\text{right}})$ connected by a bridge entity, or returns $\bot$ when the query is non-decomposable (Figure~\ref{fig:query_decomposition_prompt}).
\end{itemize}

\paragraph{Semantic coverage verifier $\mathcal{V}$.}
\begin{itemize}
    \item \textbf{Round-Trip Consistency Verification}, which checks whether the proposed split $(q_{\text{left}}, q_{\text{right}})$ jointly preserves the intent of $q$ without omission, addition, or reordering, and repairs the split when this constraint is violated (Figure~\ref{fig:round_trip_prompt}); a null \texttt{needs} field means the split passed and the original pair is kept, while a non-null field replaces it.
\end{itemize}

\paragraph{Final synthesis.}
\begin{itemize}
    \item \textbf{Final Answer}, which aggregates the resolved sub-answers $a_{\text{left}}$, $a_{\text{right}}$ and the accumulated evidence into a final response consistent with $Q$ (Figure~\ref{fig:final_answer_prompt}).
\end{itemize}

\begin{figure*}[htb]
\begin{tcolorbox}
[title = Binary Query Decomposition, colback = gray!10, colframe = black, sharp corners, boxrule=0.5mm]

You are a Multi-hop Question Decomposition Agent.
Your task is to analyze a given question and determine whether it exhibits a dependency structure that can be decomposed into \textbf{EXACTLY TWO} information needs connected by a conceptual or factual \textbf{BRIDGE}.
A \textbf{BRIDGE} is a pivot fact that must be resolved first in order to answer the final question. \\

\textbf{Typical BRIDGEs include:}
\begin{itemize}
    \item Key entities (persons, organizations, locations, dates)
    \item Important noun phrases (titles, concepts, objects)
    \item Logical or temporal relationships (cause, dependency, sequence)
    \item Explicit constraints stated in the question
\end{itemize}

\textbf{Decomposition Instructions:}
\begin{enumerate}
    \item \textbf{If the question requires an intermediate BRIDGE:} Decompose it into \textbf{EXACTLY TWO} needs:
        \begin{itemize}
            \item (N1) The pivot need that establishes the bridge
            \item (N2) The dependent need that uses the resolved bridge to reach the final answer
        \end{itemize}

    \item \textbf{If there is no intermediate BRIDGE} and the question therefore cannot be decomposed into two dependent needs: Do \textbf{NOT} decompose.
\end{enumerate}

\textbf{Rules:}
\begin{itemize}
    \item \textbf{NEVER} produce more than two needs.
    \item \textbf{NEVER} produce fewer than two needs when decomposing.
    \item \textbf{N2 MUST} logically depend on N1.
\end{itemize}

\textbf{For each need, specify:}
\begin{itemize}
    \item \texttt{id}: \texttt{N1} or \texttt{N2} only
    \item \texttt{text}: a declarative description of the information required
    \item \texttt{depends\_on}: \texttt{[]} for N1, \texttt{["N1"]} for N2
    \item \texttt{subquery}: a concise natural-language question that retrieves the required information
\end{itemize}

\textbf{Output Format:}
\begin{itemize}
    \item Always output a \textbf{JSON object}.
    \item Use \textbf{EXACTLY TWO} top-level keys: \texttt{"thought"} and \texttt{"needs"}.
    \item If the question cannot be decomposed into two dependent needs, return:
    \begin{center}
    \texttt{\{"thought": "...", "needs": null\}}
    \end{center}
\end{itemize}

Question: \texttt{\{question\}} \\

Return the results in a FLAT JSON format. \\
\textbf{DO NOT} include any explanations or notes in the output. \textbf{ONLY} return JSON.

\end{tcolorbox}
\caption{Binary Query Decomposition Prompt.}
\label{fig:query_decomposition_prompt}
\end{figure*}

\newpage
\begin{figure*}[htb]
\begin{tcolorbox}
[title = Round-Trip Consistency Verification, colback = gray!10, colframe = black, sharp corners, boxrule=0.5mm]

You are a Decomposition Verification \& Repair Agent.
Your task is to verify and, if necessary, repair a proposed \textbf{TWO-NEED} decomposition so that it faithfully represents the intent and constraints of the original multi-hop question. \\

\textbf{You will be given:}
\begin{itemize}
    \item An original multi-hop question
    \item A decomposition containing two needs: N1 and N2
\end{itemize}

\textbf{Your tasks:}
\begin{enumerate}
    \item \textbf{VERIFY:} Check whether N1 and N2, when recomposed, are equivalent to the original question and capture \emph{all} required constraints.
    \item \textbf{REPAIR (only if invalid):} If the decomposition is invalid, produce a corrected decomposition that satisfies the full intent and constraints of the original question.
\end{enumerate}

\textbf{Decomposition Rules:}
Same as those defined in the \emph{Binary Query Decomposition} prompt. \\

\textbf{For each need, specify:}
\begin{itemize}
    \item \texttt{id}: \texttt{N1} or \texttt{N2} only
    \item \texttt{text}: a declarative description of the information required
    \item \texttt{depends\_on}: \texttt{[]} for N1, \texttt{["N1"]} for N2
    \item \texttt{subquery}: a concise natural-language question that retrieves the required information
\end{itemize}

\textbf{Output Format:}
\begin{itemize}
    \item Always output a \textbf{JSON object} with \textbf{EXACTLY TWO} top-level keys: \texttt{"thought"} and \texttt{"needs"}.
    \item If the decomposition is \textbf{valid} and already satisfies all constraints:
    \begin{itemize}
        \item Set \texttt{"needs"} to \texttt{null}.
    \end{itemize}
    \item If the decomposition is \textbf{invalid}:
    \begin{itemize}
        \item Set \texttt{"needs"} to the corrected two-need decomposition.
    \end{itemize}
\end{itemize}

Question: \texttt{\{question\}} \\

Decomposition: \texttt{\{decomposition\}} \\

Return the results in a FLAT JSON format. \\
\textbf{DO NOT} include any explanations or notes in the output. \textbf{ONLY} return JSON.

\end{tcolorbox}
\caption{Round-Trip Consistency Verification Prompt.}
\label{fig:round_trip_prompt}
\end{figure*}

\newpage
\begin{figure*}[htb]
\begin{tcolorbox}
[title = Granularity Assessment (Root version), colback = gray!10, colframe = black, sharp corners, boxrule=0.5mm]

You are an advanced reading comprehension assistant. \\ 
Your task is to analyze text passages and corresponding questions meticulously.\\

Always respond as a JSON object with the following structure: \\
\texttt{
\{ "thought": "<methodically break down the reasoning process, illustrating how you arrive at conclusions>", \\ "answer": "<a concise and definitive answer string>" \}
}

\begin{itemize}
    \item If there is an answer, extract the answer span from the text passages.
    \item If there is no answer, respond with
    \texttt{\{ "thought": "<methodically break down the reasoning process, illustrating how you arrive at conclusions>", "answer": null \}}
\end{itemize}

Question: \texttt{\{question\}} \\

Documents: \texttt{\{documents\}} \\

Return the results in a FLAT JSON format. \\
\textbf{DO NOT} include any explanations or notes in the output. \textbf{ONLY} return JSON.

\end{tcolorbox}
\caption{Granularity Assessment Prompt (Root Version).}
\label{fig:granularity_assessment_root_prompt}
\end{figure*}

\newpage
\begin{figure*}[htb]
\begin{tcolorbox}
[title = Granularity Assessment (Recursive version), colback = gray!10, colframe = black, sharp corners, boxrule=0.5mm]
You are a QA assistant. \\

Context: 
\begin{itemize}
    \item The ORIGINAL QUESTION is being solved through multiple intermediate steps.
    \item Some intermediate facts have ALREADY been resolved in previous steps.
    \item These resolved facts are provided in the Current Document Context.
    \item The current SUBQUESTION is derived from that context and represents ONLY ONE intermediate step.
    \item There may be further steps after this one.
    \item The ORIGINAL QUESTION and the Current Document Context are provided ONLY to help you interpret the SUBQUESTION.
\end{itemize}

Your task:
\begin{itemize}
    \item Answer ONLY the given SUBQUESTION.
    \item Do NOT attempt to answer the ORIGINAL QUESTION.
    \item You may use the ORIGINAL QUESTION to understand how the result will be used.
    \item If the SUBQUESTION is NOT the final step toward answering the ORIGINAL QUESTION, return the answer that will be most useful for subsequent reasoning steps.
    \item If the SUBQUESTION IS the final step, return the answer directly.
\end{itemize}

Always respond as a JSON object with the following structure: \\
\texttt{
\{ "thought": "<methodically break down the reasoning process, illustrating how you arrive at conclusions>", \\ "answer": "<a concise and definitive answer string>" \}
}

\begin{itemize}
    \item If there are MULTIPLE valid answers to the SUBQUESTION, return ALL of them.
    \item If there is no answer, respond with
    \texttt{\{ "thought": "<methodically break down the reasoning process, illustrating how you arrive at conclusions>", "answer": null \}}
\end{itemize}

Original Question: \texttt{\{original question\}} \\

Current Document Context: \texttt{\{retrieved documents so far\}} \\

Subquestion: \texttt{\{sub question\}} \\

Documents: \texttt{\{documents\}} \\

Return the results in a FLAT JSON format. \\
\textbf{DO NOT} include any explanations or notes in the output. \textbf{ONLY} return JSON.

\end{tcolorbox}
\caption{Granularity Assessment Prompt (Recursive Version).}
\label{fig:granularity_assessment_recursive_prompt}
\end{figure*}

\newpage
\begin{figure*}[htb]
\begin{tcolorbox}
[title = Subquery Refinement, colback = gray!10, colframe = black, sharp corners, boxrule=0.5mm]
You are a subquery planner for multi-hop QA. \\

You are given:
\begin{itemize}
    \item the original question Q
    \item the current target need
    \item the retrieved documents accumulated so far
\end{itemize}

Your job:
\begin{itemize}
    \item Construct a focused, concise text query $q_{\text{next}}$ that will help satisfy the outstanding needs.
    \item The query should incorporate any necessary entities from known facts in the retrieved docs.
\end{itemize}

Output format:
\begin{itemize}
    \item Return ONLY a JSON object: \texttt{\{"query": "..."\}}
    \item No explanations.
\end{itemize}

Original Question: \texttt{\{original question\}} \\

Target need: \texttt{\{subquery\}} \\

Retrieved documents: \texttt{\{documents\}} \\

Return the results in a FLAT JSON format. \\
\textbf{DO NOT} include any explanations or notes in the output. \textbf{ONLY} return JSON.

\end{tcolorbox}
\caption{Subquery Refinement Prompt.}
\label{fig:subquery_refinement_prompt}
\end{figure*}

\newpage
\begin{figure*}[htb]
\begin{tcolorbox}
[title = Final Answer, colback = gray!10, colframe = black, sharp corners, boxrule=0.5mm]
You are an advanced reading comprehension assistant. \\

You are given:
\begin{itemize}
    \item Question
    \item A history of decomposed information needs and the answers found for each need
    \item The retrieved documents accumulated during multi-hop retrieval
\end{itemize}

Your task: Answer the Question using the provided history and retrieved documents. \\

Always respond as a JSON object with the following structure: \\
\texttt{
\{ "thought": "<methodically break down the reasoning process, illustrating how you arrive at conclusions>", \\ "answer": "<final answer>" \}
} \\

Original Question: \texttt{\{original question\}} \\

History (needs and answers): \texttt{\{history\}} \\

Retrieved documents: \texttt{\{documents\}} \\

Return the results in a FLAT JSON format. \\
\textbf{DO NOT} include any explanations or notes in the output. \textbf{ONLY} return JSON.

\end{tcolorbox}
\caption{Final Answer Prompt.}
\label{fig:final_answer_prompt}
\end{figure*}

\newif\ifchecklist
\checklistfalse
\ifchecklist
  \clearpage
  \input{checklist.tex}
\fi

\end{document}